\documentclass[journal]{IEEEtran}
\usepackage{color}
\newcommand{\sunnie}[1]{{\color{black} #1}}

\usepackage{cite}
\providecommand{\shortcite}[1]{\cite{#1}}
\ifCLASSINFOpdf
   \usepackage[pdftex]{graphicx}
\else
\fi
\usepackage{booktabs} % For formal tables
\usepackage{subfigure}
\usepackage{multirow}
\usepackage{url}
\begin{document}
\title{Moir\'{e} Photo Restoration Using Multiresolution Convolutional Neural Networks}

\author{Yujing~Sun,
        Yizhou~Yu,
       Wenping~Wang
\IEEEcompsocitemizethanks{\IEEEcompsocthanksitem
 Department of Computer Science, The University of Hong Kong, Pokfulam Road, Hong Kong.
 E-mail: \{yjsun, wenping\}@cs.hku.hk, yizhouy@acm.org
%\protect\\
}

%\thanks{This paper has supplementary downloadable material available at http://ieeexplore.ieee.org. The material includes more experimental results. Contact yjsun@cs.hku.hk for further questions about this work.}
}


\markboth{Journal of \LaTeX\ Class Files,~Vol.~14, No.~8, August~2015}%
{Shell \MakeLowercase{\textit{et al.}}: Bare Demo of IEEEtran.cls for IEEE Journals}



\maketitle

\begin{abstract}
Digital cameras and mobile phones enable us to conveniently record precious moments. While digital image quality is constantly being improved, taking high-quality photos of digital screens still remains challenging because the photos are often contaminated with moir\'{e} patterns, a result of the interference between the pixel grids of the camera sensor and the device screen. Moir\'{e} patterns can severely damage the visual quality of photos. 
\sunnie{However, few studies have aimed to solve this problem.}
In this paper, we introduce a novel multiresolution fully convolutional network for automatically removing moir\'{e} patterns from photos. Since a moir\'{e} pattern spans over a wide range of frequencies, our proposed network performs a nonlinear multiresolution analysis of the input image before computing how to cancel moir\'{e} artefacts within every frequency band. We also create a large-scale benchmark dataset with $100,000^+$ image pairs for investigating and evaluating moir\'{e} pattern removal algorithms. Our network achieves state-of-the-art performance on this dataset in comparison to existing learning architectures for image restoration problems.
\end{abstract}


% \fi
\IEEEpeerreviewmaketitle

\maketitle

\section{Introduction}
\IEEEPARstart{N}{owadays}, digital cameras and mobile phones play a significant role in people's lives. They enable us to easily record any precious moments that are interesting or meaningful. There exist many occasions when people would like to capture digital screens. Such occasions include taking photos of visual contents on a screen, or shooting scenes involving digital monitors. While image quality is constantly being improved, taking high-quality photos of digital screens still remains challenging. Such photos are often contaminated with moir\'{e} patterns (Fig. ~\ref{fig:moireDemo}).

A moir\'{e} pattern in the photo of a screen is the result of the interference between the pixel grids of the camera sensor and the device screen. It can appear as stripes, ripples, or curves of intensity and colour diversifications superimposed onto the photo. The moir\'{e} pattern can vary dramatically due to a slight change in shooting distance or camera orientation. This moir\'{e} artefact severely damages the visual quality of the photo. There is a large demand for post-processing techniques capable of removing such artefacts. In this paper, we call images of digital screens taken with digital devices \textit{moir\'{e} photos}.

\begin{figure}[t]
%--
\begin{center}
\subfigure{
\includegraphics[width = 0.44\linewidth]{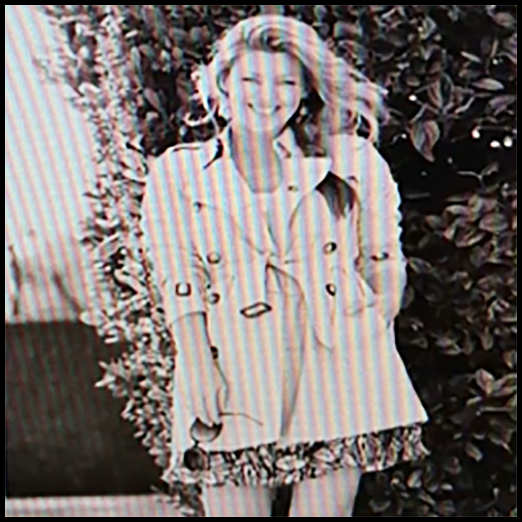}}
\subfigure{
\includegraphics[width = 0.44\linewidth]{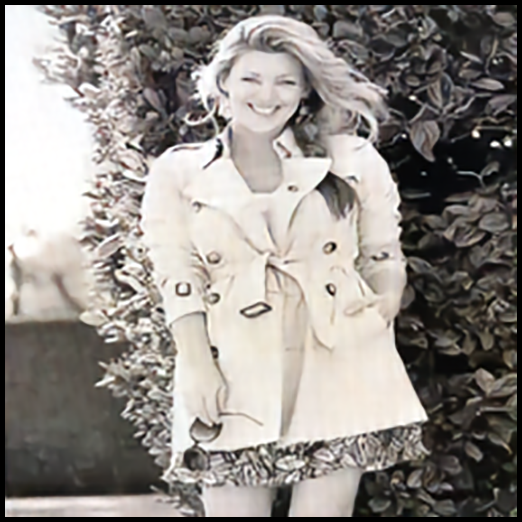}}
\end{center}
\caption{Given an image damaged by moir\'{e} patterns, our proposed network can remove the moir\'{e} artefacts automatically.}

\label{fig:teaser}
\end{figure}

It is particularly challenging to remove moir\'{e} patterns in photos, which are mixed with original image signals across a wide range in both spatial and frequency domains. A moir\'{e} pattern typically covers an entire image. The colour or thickness of the stripes or ripples in such patterns not only changes from image to image, but also is spatially varying within the same image. Thus, a moir\'{e} pattern could occupy a high-frequency range in one image region, but a low-frequency range in another region. Due to the complexity of moir\'{e} patterns in photos, little research has been dedicated to moir\'{e} pattern removal. Conventional image denoising~\cite{chen2013fast}, and texture removal techniques~\cite{xu2012structure,cho2014bilateral} are not well suited for this problem because these techniques typically assume noises and textures occupy a higher-frequency band than true image structures.

On the other hand, convolutional neural networks are leading a revolution in computer vision and image processing.
After successes in image classification and recognition~\cite{simonyan2014very,szegedy2015going}, they have also been proven highly effective in low-level vision and image processing tasks, including image super-resolution~\cite{dong2014learning,kim2016accurate}, demosaicking~\cite{demosaick2016deep}, denoising~\cite{zhang2017beyond}, and restoration~\cite{zhang2017learning}.

In this paper, we introduce a novel multiresolution fully convolutional neural network for automatically removing moir\'{e} patterns from photos. Since a moir\'{e} pattern spans over a wide range of frequencies, to make the problem more tractable, our network first converts an input image into multiple feature maps at various different resolutions, which include different levels of details. Each feature map is then fed into a stack of cascaded convolutional layers that maintain the same input and output resolutions. These layers are responsible for the core task of canceling the moir\'{e} effect associated with a specific frequency band. The computed components at different resolutions are finally upsampled to the input resolution and fused together as the final output image.

To train and test our multiresolution network, we also create a dataset of 135,000 image pairs, each containing an image contaminated with moir\'{e} patterns and its corresponding uncontaminated reference image. The reference images are taken from the ImageNet dataset. The contaminated images have a wide variety of moir\'{e} effects. They are obtained by taking photos of reference images displayed on a computer screen using a mobile phone. To our knowledge, this is the first large-scale dataset for research on moir\'{e} pattern removal. The proposed network achieves state-of-the-art performance on this dataset, compared with existing learning architectures for image restoration problems.

We summarise our contributions in this paper as follows.
\begin{itemize}
\item[1] We present a novel and highly effective learning architecture for restoring images contaminated with moir\'{e} patterns.
\item[2] We also create the first large-scale benchmark dataset for moir\'{e} pattern removal. This dataset contains $100,000^+$ image pairs, and will be publicly released for research and evaluation.
\end{itemize}

\section{Background and Related Work}
\subsection{The Moir\'{e} Effect}
\begin{figure}[t]
\begin{center}
 \includegraphics[width=0.7\linewidth]{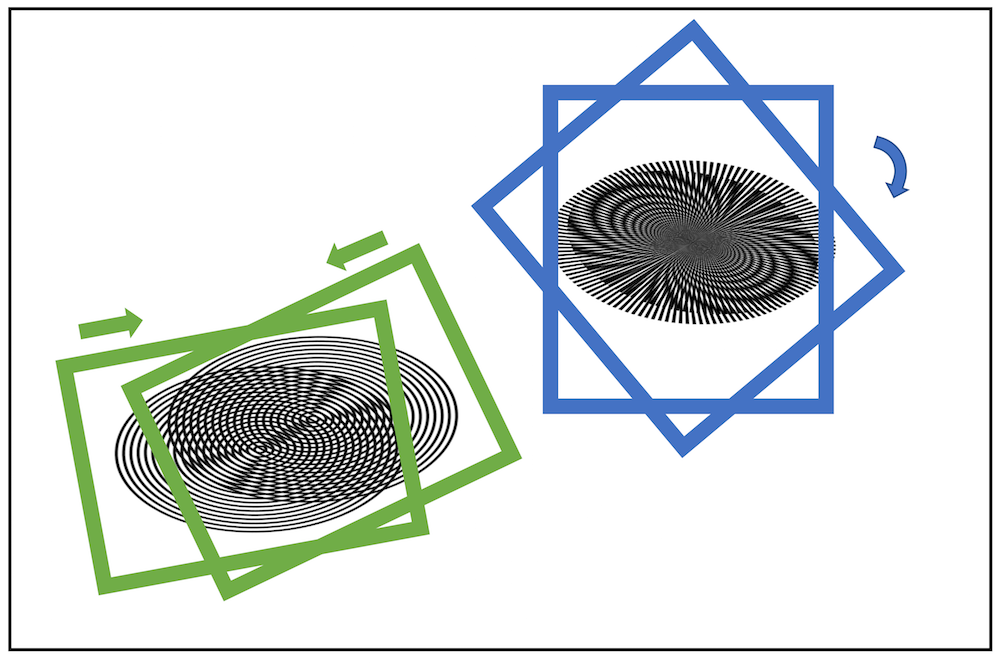}
\end{center}
 \caption{The mechanism underlying a general moir\'{e} pattern. The changing misalignment between two repetitive patterns produces varying moir\'{e} patterns.}
\label{fig:generalMoireDemo}
\end{figure}

When two similar, repetitive patterns of lines, circles, or dots overlap with imperfect alignment, a new dynamic pattern appears. This new pattern is called the moir\'{e} pattern, which can involve multiple colours. A moir\'{e} pattern changes the shape and frequency of its elements when the two original patterns move relative to each other (Fig.~\ref{fig:generalMoireDemo}).

Moir\'{e} patterns are large-scale interference patterns. For such interference patterns to occur, the two original patterns must not be completely aligned. Moir\'{e} patterns magnify misalignments. The slightest misalignment between the two original patterns could give rise to a large-scale, easily visible moir\'{e} pattern. As the degree of misalignment increases, the frequency of the moir\'{e} pattern may also increase.

Moir\'{e} patterns often occur as an artefact of images generated by digital imaging or computer graphics techniques, such as when scanning a printed halftone picture or rendering a checkerboard pattern that extends toward the horizon~\cite{WinNT}. The latter is also a case of aliasing due to undersampling a fine regular pattern.

\paragraph{Moir\'{e} Photos}
Photographs of a computer or TV screen taken with a digital camera often exhibit moir\'{e} patterns.
Examples are shown in Fig.~\ref{fig:moireDemo}. This is because a screen consists of a grid of pixels while the camera sensor is another grid of pixels. When one grid is mapped to another grid, pixels in these two grids do not line up exactly, giving rise to moir\'{e} patterns.

Similar to the formation of general moir\'{e} patterns, when the relative position between a screen and a digital camera changes, the moir\'{e} pattern in the image can change dramatically. It can be 1) of various types: stripes, dots or waves, 2) of various scales, 3) of various levels of intensity, 4) anisotropic or isotropic, and 5) uniform or non-uniform. Removing such moir\'{e} patterns with diverse properties is a challenging problem.

The occurrence of moir\'{e} patterns in photographs of computer or TV screens does not indicate a defect in the screen but is a result of a practical limitation in display technology. In order to completely eliminate moir\'{e} patterns, the dot or stripe pitch on the screen would have to be significantly smaller than the size of a pixel in the camera, which is generally not possible~\cite{keohi}.

%--------------------------------------------------------------------------------------------------------------------------------------------------

%FIGURE moireDemo -------------------------------------------------------------------------
%------------------------------------------------------------------------

%----------------------------------------------------------------------------------------------------------------------------------------------------
\subsection{Related Work}
\textbf{Moir\'{e} Pattern Removal} Several methods have been proposed to remove different types of moir\'{e} patterns. Sidorov and Kokaram~\cite{sidorov2002suppression} presented a spectral model to suppress moir\'{e} patterns in film-to-video transfer using telecine devices. However, the moir\'{e} patterns they deal with are monotonous and monochrome. Thus, their method is unsuitable for eliminating the moir\'{e} patterns in our context. Observing that moir\'{e} patterns on textures are dissimilar while a texture is locally well-patterned, Liu et al.~\cite{liu2015moire} proposed a low-rank and sparse matrix decomposition method to remove moir\'{e} patterns on high-frequency textures. Because our moir\'{e} patterns occur on high-frequency textures as well as on low-frequency structures, the method in \cite{liu2015moire} is unable to solve our problem. Taking advantage of frequency domain statistics, Sur and Gr\'{e}diac~\cite{sur2015automated} proposed to remove quasi-periodic noise. Different from our moir\'{e} patterns, quasi-periodic noise is simple and regular. Due to the complexity of our moir\'{e} patterns, aforementioned methods cannot remove the artefacts well while preserving the original image appearance.

\textbf{Image Descreening} \quad In order to print continuous tone images, most electrophotographic printers take advantage of the halftoning techniques, which rely on local dot patterns to approximate continuous tones. Scanned copies of such printed images are commonly corrupted with screen-like high-frequency artefacts (moir\'{e} effect), exhibiting low aesthetic quality. Image descreening aims at reconstructing high-quality images from scanned versions of images printed using halftoning (such as scanned books), and has been well studied in the past decades. Various methods have been proposed, such as printer-end algorithms~\cite{descreen2001adaptive,descreen2004amfm}, image smoothing techniques~\cite{descreen1995smooth}, learning based methods~\cite{descreen2007training,descreen2004GACNN}, and advanced filters~\cite{descreen1998wavelet,descreen2010hardware,descreen2014adaptive}.
Specialised methods have been proposed to process a specific subset of images, such as paper checks~\cite{ok2017paper}. Shou and Lin~\shortcite{descreen2004GACNN} descreened images on the basis of a learning based pattern classification process. They found that it is sufficient to consider two classes of moir\'{e} patterns to produce satisfactory results. The reason is that halftoning typically involves binary colours, and that the viewing distance and angle during scanning are almost fixed. Such constraints make moir\'{e} patterns in the descreening problem regular, uniform, and local. Therefore, existing image descreening techniques are inadequate to deal with our complex moir\'{e} patterns.

\textbf{Texture Removal} \quad Since moir\'{e} patterns in photos often have high-frequency and repetitive components, texture removal algorithms are a class of relevant techniques. Xu {\em et al.}~\shortcite{xu2012structure} introduced relative total variation to describe and identify textures. Karacan {\em et al.}~\shortcite{karacan2013structure} took advantage of region covariances to separate texture from image structure. Ono {\em et al.}~\shortcite{ono2014cartoon} utilised block-wise low-rank texture characterisation to decompose images into texture and structure components. Cho {\em et al.}~\shortcite{cho2014bilateral} combined the bilateral filter with a ``patch shift" texture range kernel to achieve a similar goal. Sun {\em et al.}~\shortcite{sun2017image} took advantage of $l_0$ norm to retrieve structures from textured images. Ham {\em et al.}~\shortcite{ham2015} performed texture removal through image filtering with joint static and dynamic guidance. State-of-the-art methods define a variety of local filters to remove high-frequency textures. However, moir\'{e} patterns in photos are not merely high-frequency artefacts but span a wide range of frequencies. In addition, moir\'{e} patterns also introduce colour distortions, which existing texture removal algorithms would not be able to remove.

\textbf{Image Restoration} Image restoration problems aim at removing noises or reconstructing high-frequency details. Recently, learning techniques have been successfully applied to image restoration tasks, including image super-resolution~\cite{dong2014learning,kim2016accurate,zhang2017learning}, denoising~\cite{zhang2017beyond,zhang2017learning}, and
deblurring~\cite{nah2016deep,zhang2017learning}. These learning based methods have achieved state-of-the-art performance in image quality improvement. The problem we aim to solve in this paper can be considered as a special image restoration problem as well since it attempts to reconstruct the uncontaminated image by removing moir\'{e} artefacts. However, different from the uniformly distributed noises in the denoising task and the missing high-frequency details in the super-resolution task, the moir\'{e} patterns in our problem can be anisotropic and non-uniform, and exhibit features across a wide range of frequencies. The models employed in traditional image restoration tasks are not specifically tailored for our problem and can only achieve suboptimal performance. Most recently, Gharbi {\em et al.}~\shortcite{demosaick2016deep} presented a learning-based method to demosaic and denoise images. However, demosaicking is also limited to removing high-frequency artefacts only.

%FIGURE modelDemo -------------------------------------------------------------------------
\begin{figure*}[t]
\begin{center}
 \includegraphics[width=0.85\linewidth]{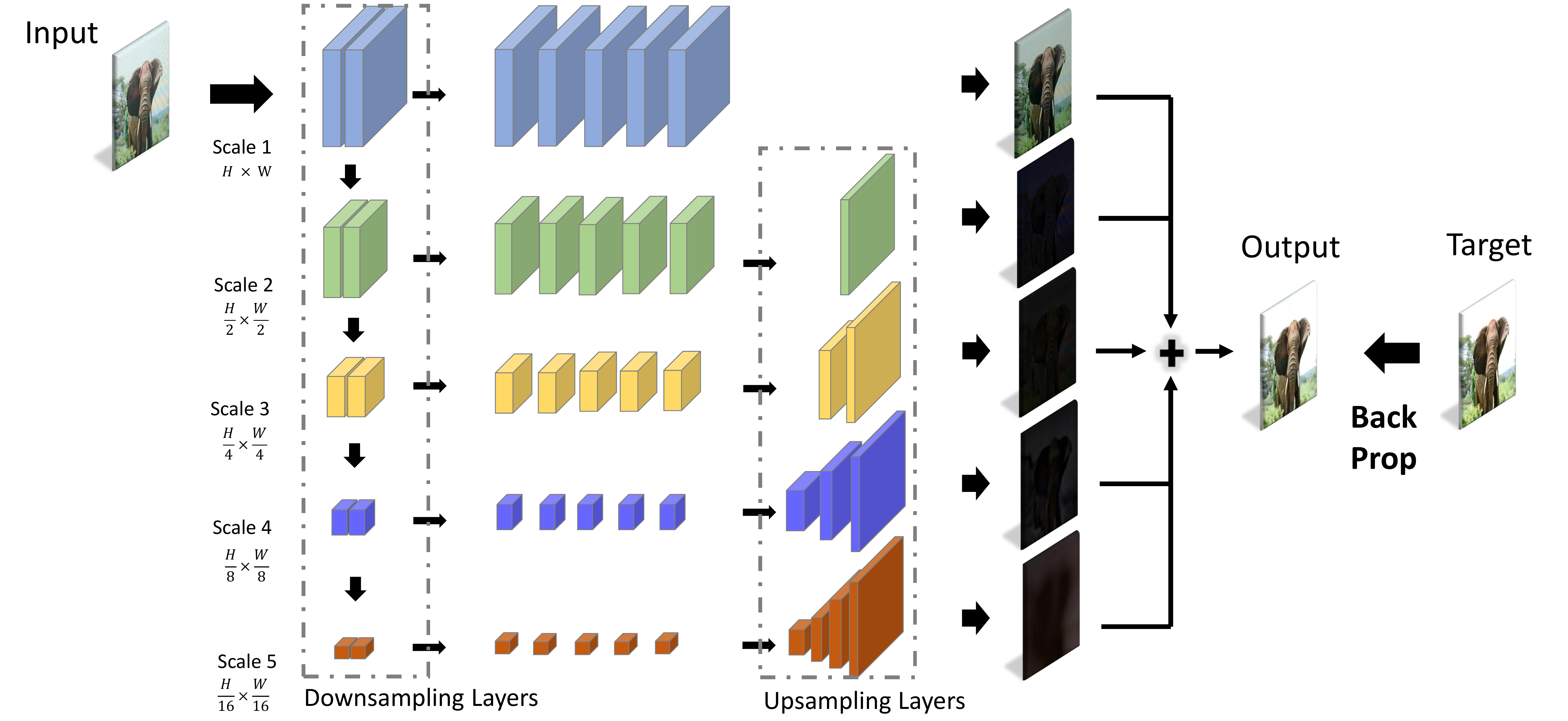}
\end{center}

\begin{center}
\hspace{-.5em}
\subfigure[Input]{
 \includegraphics[width=0.175\linewidth]{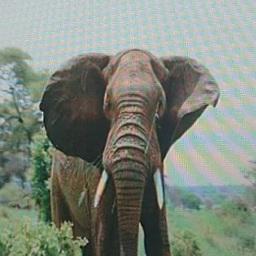}}
 \hspace{-.6em}
\subfigure[Finest Scale]{
 \includegraphics[width=0.175\linewidth]{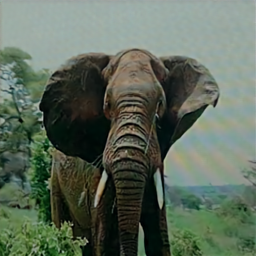}}
 \hspace{-.6em}
 \subfigure[Scales 2 to 5]{
 \includegraphics[width=0.35\linewidth]{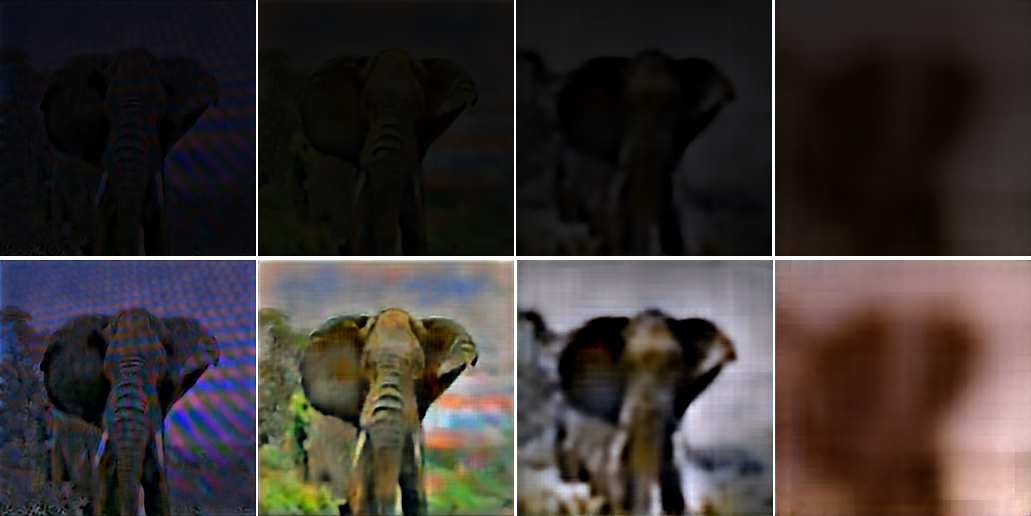}}
 \hspace{-.6em}
 \subfigure[Output]{
 \includegraphics[width=0.175\linewidth]{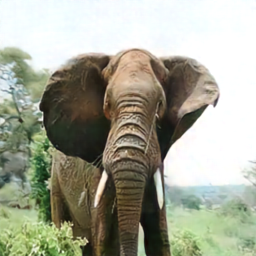}}
\end{center}
 \caption{The architecture of our multiresolution fully convolutional network. The top row in (c) shows intermediate images produced from the second to fifth network branch, and the bottom row shows the same images with amplified intensity.}
\label{fig:modeldemo}
\end{figure*}
%------------------------------------------------------------------------

\section{Multiresolution Deep CNN for Moir\'{e} Pattern Removal}
By considering problem complexity, we choose CNNs to remove moir\'{e} patterns in photographs due to their recent impressive performance on image restoration tasks.  
In this section, we present a multiresolution fully convolutional neural network to tackle the problem. It exploits intrinsic correlations between moir\'{e} patterns and image components at different levels of a multiresolution pyramid. The training process of our network jointly optimises all parameters to minimize the loss function. As shown in Fig.~\ref{fig:teaser}, once trained, our network can automatically remove moir\'{e} patterns in contaminated images.

\subsection{Network Architecture}
Our network architecture is outlined in Fig.~\ref{fig:modeldemo}, which includes multiple parallel branches at different resolutions. The branch at the top processes feature maps at the original resolution of the input image while other branches process coarser and coarser feature maps. The first two convolutional layers in each branch form a group and are responsible for downsampling the feature maps from the immediate higher-level branch by half if there is such a higher-level branch. Therefore the feature maps generated after the first two convolutional layers at all branches can be stacked together to form an upside-down pyramid, where any feature map has half of the resolution of the feature map at the next higher level. Interestingly, in contrast to traditional image pyramids computed using linear filters, our pyramid is computed using nonlinear ``filters" (i.e. convolutional kernels + nonlinear activation functions). By converting the input image into multiple feature maps at various different resolutions, we aim to expose different levels of details in the input image.

Inside each branch, the output feature maps from the first two layers are fed into a sequence of cascaded convolutional layers. These convolutional layers maintain the same input and output resolutions, and do not perform any downsampling or pooling operations. They are responsible for the core task of canceling the moir\'{e} effect associated with the specific frequency band of that branch. Even with the above multiresolution analysis, this is still a hard task that involves sophisticated nonlinear transforms. Therefore, we place multiple convolutional layers (typically 5) each with $3 \times 3$ kernels and 64 channels in this sequence.

To assemble the transformed results from all parallel branches together into a complete output image, we still need to increase the resolution of the feature map generated from the cascaded convolutional layers to the original resolution of the input image within each branch except for the first one. In the $i$-th branch from the top, we use a set of $i-1$ deconvolutional layers to achieve this goal. Each deconvolutional layer doubles the input resolution. There is an extra convolutional layer following the deconvolutional layers within each branch. This extra layer generates a feature map with 3 channels only. This feature map essentially cancels the component of the moir\'{e} pattern (in the input image) associated with the frequency band of that branch. At the end, the final 3-channel feature maps from all branches are simply summed together to produce the final output image with the moir\'{e} pattern removed.

In our network, whenever there is a need to reduce the resolution of a feature map by half, we use a kernel stride 2 instead of a pooling layer. Each layer is followed by a rectified linear unit (ReLU) and we pad zeros to ensure that the output of each layer is of desired size. The detailed configurations of the first two layers and last layers within all branches are given in Table.~\ref{table:downlayerspec} and Table.~\ref{table:uplayerspec}, respectively.

\begin{table}[thb!]
\begin{minipage}[t]{.5\linewidth}
\caption{Downsampling Layers}
\centering
\label{table:downlayerspec}
\resizebox{.9\textwidth}{!}{ \begin{tabular}{@{}cccc@{}}
\toprule
Scale & Kernel  & Stride & Channels\\ \midrule
 1     & 3x3    		&1x1       &32 \\
 1     & 3x3      		&1x1       &32 \\
\hline
 2     & 3x3      & 2x2      &32\\
 2     & 3x3      & 1x1      &64\\
\hline
 3     & 3x3     & 2x2      &64\\
 3     & 3x3      & 1x1      &64\\
\hline
 4     & 3x3      & 2x2      &64 \\
 4     & 3x3      & 1x1      &64\\
\hline
 5     & 3x3      & 2x2      &64 \\
 5     & 3x3      & 1x1      &64 \\ \bottomrule
\end{tabular}}
\end{minipage}%
\begin{minipage}[t]{.5\linewidth}
\centering
\caption{Upsampling Layers}
\label{table:uplayerspec}
\resizebox{1.0\textwidth}{!}{ \begin{tabular}{@{}ccccc@{}}
\toprule
Scale & Type & Kernel & Stride & Channels     \\ \midrule
1     &conv	& 3x3                		& 1x1      & 3       \\
\hline
2     &deconv	& 4x4                		& 2x2      & 32      \\
     &conv		& 3x3                		& 1x1      & 3     \\
\hline
3     &deconv	& 4x4                		& 2x2      & 64    \\
     &deconv	& 4x4                		& 2x2      & 32       \\
     &conv		& 3x3                		& 1x1      & 3        \\
\hline
4    &deconv	& 4x4                		& 2x2        & 64      \\
     &deconv	& 4x4                		& 2x2      & 32       \\
     &deconv	& 4x4                		& 2x2      & 32        \\
    &conv		& 3x3                		& 1x1      & 3       \\
\hline
5     &deconv	& 4x4                		& 2x2      & 64      \\
    &deconv	& 4x4                		& 2x2      & 32     \\
     &deconv	& 4x4                		& 2x2      & 32      \\
     &deconv	& 4x4                		& 2x2      & 32       \\
     &conv		& 3x3                		& 1x1      & 3    \\ \bottomrule
\end{tabular}}
\end{minipage} 

\end{table}

\paragraph{Remarks.} Our deep network is designed on the basis of the key characteristics of moir\'{e} patterns, which exhibit features across a wide range of frequencies. A moir\'{e} pattern is typically spatially varying and spreads over an entire image. If a network deals with fine-scale features only, low-frequency components of the moir\'{e} pattern cannot be removed; if it deals with coarse-scale features only, high-frequency features of the moir\'{e} pattern cannot be removed. For these reasons, we perform a multiresolution analysis of the input image and remove the component of the moir\'{e} pattern within every frequency band separately.

In Fig.~\ref{fig:modeldemo}, we illustrate how our network removes a moir\'{e} pattern from a contaminated image. The network branch for the original resolution (the finest scale) plays a dominant role because pixel colours in the final output image mostly come from this branch. We can see that moir\'{e} artefacts have not been completely removed in the 3-channel feature map produced from the last layer of the top branch (Fig.~\ref{fig:modeldemo}(b)) though such artefacts have become much weaker than those in the original input (Fig.~\ref{fig:modeldemo}(a)). Network branches for other coarser resolutions play a supporting role. The last layer of each coarser-resolution branch produces an image that aims to cancel the remaining moir\'{e} pattern (in the image produced from the last layer of the top branch) which falls into its frequency band (Fig.~\ref{fig:modeldemo}(c)). When images from all the branches are summed together, the remaining artefacts in the image from the top branch can be successfully eliminated (Fig.~\ref{fig:modeldemo}(d)).

\subsection{Network Training}
We train our deep network using a dataset of images, $D = \{ (I_i, O_i) \} $, where $I_i$ is an image contaminated with a moir\'{e} pattern and $O_i$ is its corresponding ground-truth uncontaminated image.  The training process solves for weights $\mathbf{w}$ and biases $\mathbf{b}$ in our network via minimising the following $l_2$ loss defined on image patches of size $p\times p$ from the training set $D$ in an end-to-end fashion:
\begin{equation}
L(\{ \mathbf{w}, \mathbf{b} \}) = \frac{1}{N} \sum_{i = 1}^N ||S_{i} - T_{i}||^2,
\label{eq:loss}
\end{equation}
where $N$ is the total number of image patch pairs and $(S_i, T_i)$ is a pair of patches. 

%--------------------------------------------------------------------------
\section{Dataset}

\begin{figure}[b]
\begin{center}
\subfigure{
\includegraphics[width=0.295\linewidth]{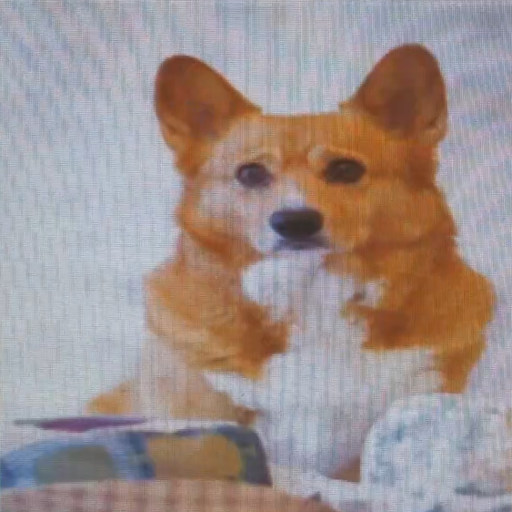}}
\hspace{-.8em}
\subfigure{
 \includegraphics[width=0.295\linewidth]{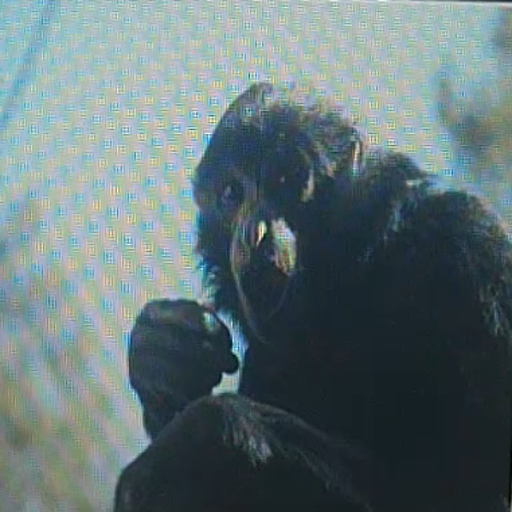}}
 \hspace{-.8em}
\subfigure{
 \includegraphics[width=0.295\linewidth]{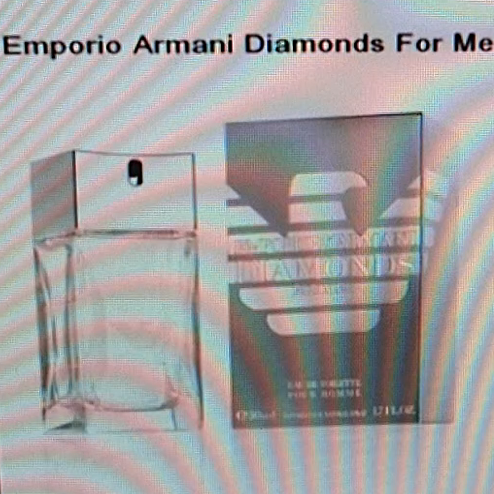}}
 \hspace{-.8em}
\end{center}
\addtocounter{subfigure}{-3}
\vspace{-1.5em}
\begin{center}
\subfigure{
 \includegraphics[width=0.295\linewidth]{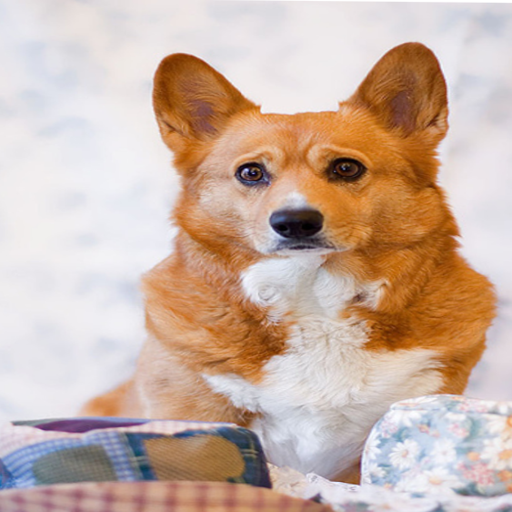}}
 \hspace{-.8em}
 \subfigure{
 \includegraphics[width=0.295\linewidth]{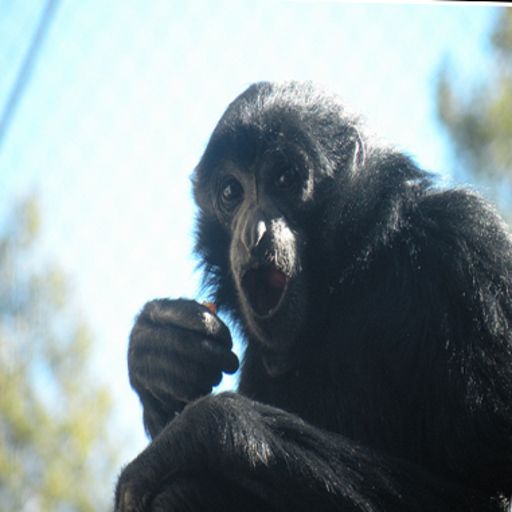}}
 \hspace{-.8em}
 \subfigure{
 \includegraphics[width=0.295\linewidth]{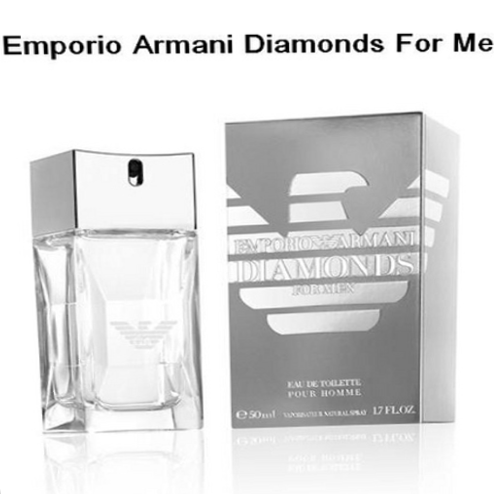}}
\end{center}
 \caption{Examples of image pairs from our dataset. From left to right: images are contaminated by stripe, dot and curved moir\'{e} patterns respectively.}
\label{fig:moireDemo}
\end{figure}

\begin{figure}[t]
\begin{center}
\subfigure[Reference image (T) ]{
\includegraphics[width = 0.3\linewidth]{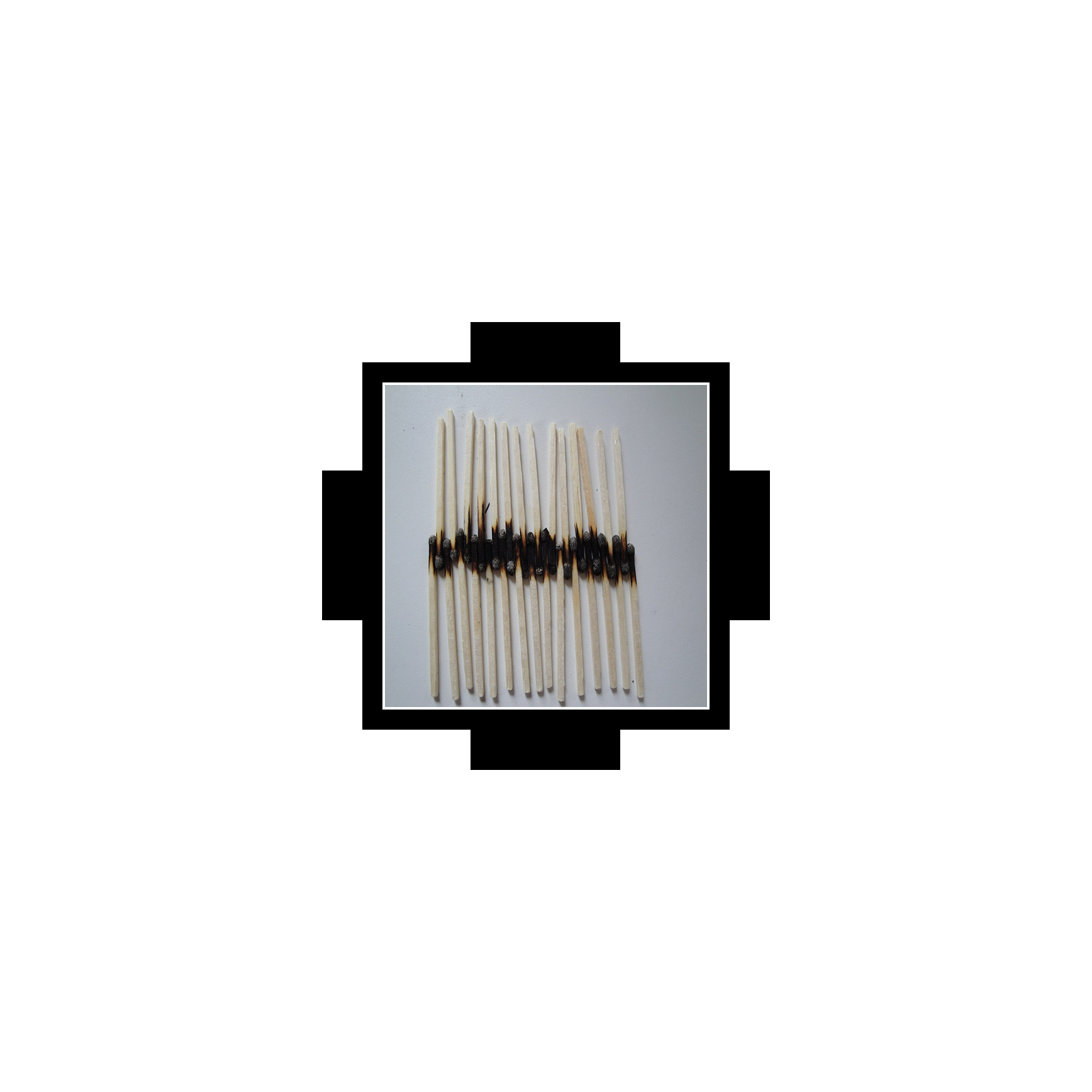}}
\subfigure[Sorted Corners of T ]{
\includegraphics[width = 0.3\linewidth]{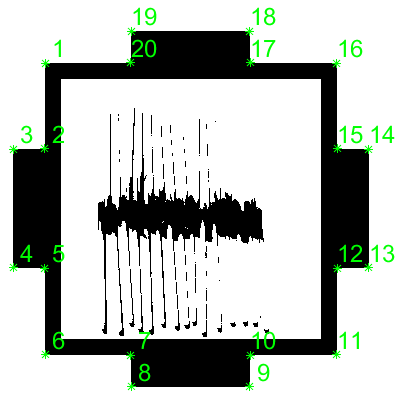}}
\subfigure[Registered T ]{
\includegraphics[width = 0.3\linewidth]{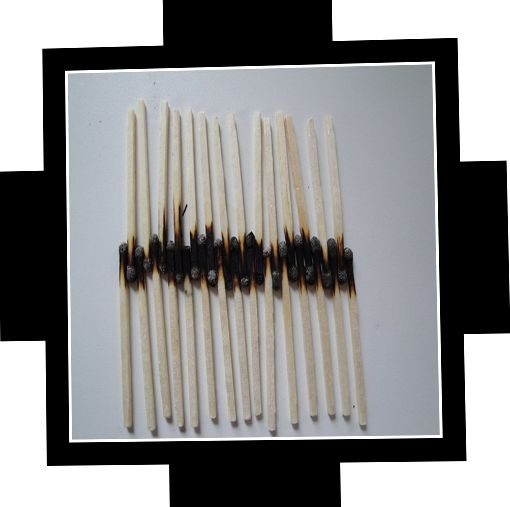}}
\end{center}
\begin{center}
\subfigure[Moir\'{e} Photo (S)]{
\includegraphics[width = 0.3\linewidth]{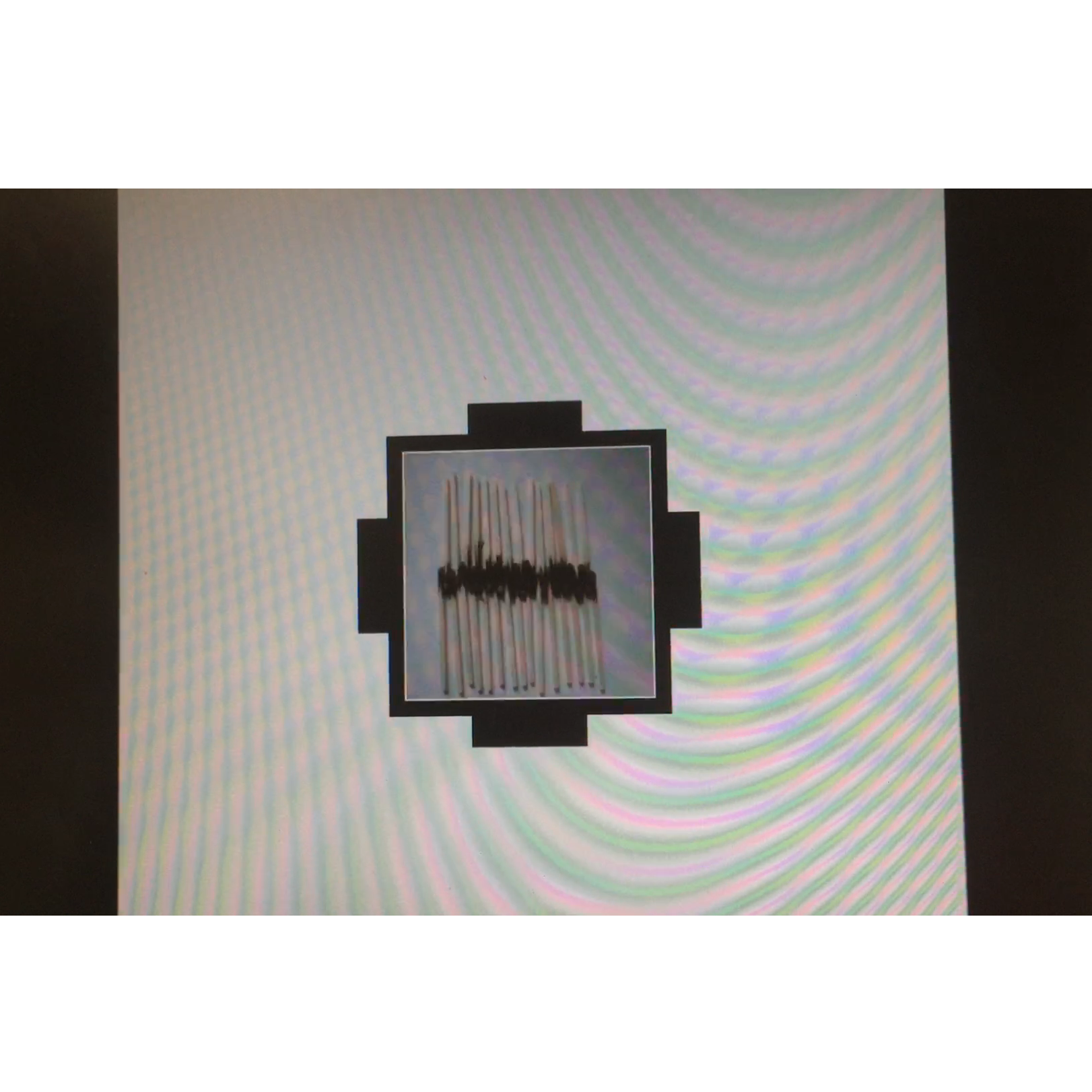}}
\hspace{-0.5em}
\subfigure[Sorted Corners of S]{
\includegraphics[width = 0.3\linewidth]{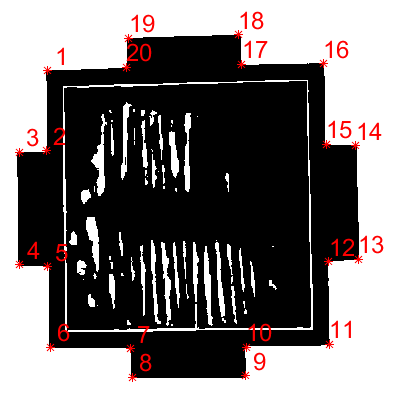}}
\hspace{-0.5em}
\subfigure[Registered S]{
\includegraphics[width = 0.3\linewidth]{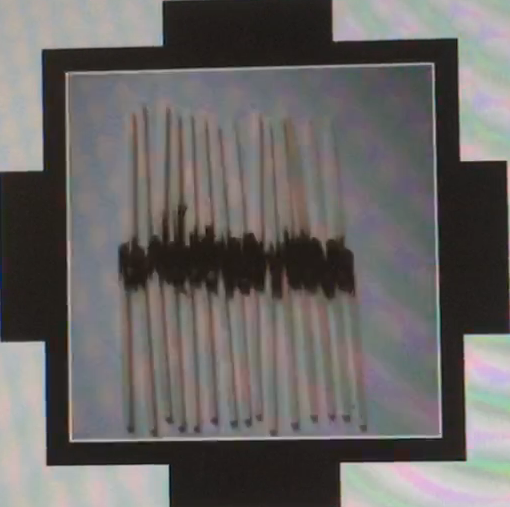}}
\end{center}
\caption{Image Acquisition. }
\label{fig:datacollection}
\end{figure}

We create a benchmark of $135,000$ image pairs, each containing an image contaminated with a moir\'{e} pattern and its corresponding uncontaminated reference image. The contaminated images have a wide variety of moir\'{e} effects (Fig. ~\ref{fig:moireDemo}).
% that are either anisotropic or isotropic, striped or dotted, and strong or weak. 
The uncontaminated reference images in our benchmark come from the 100,000 validation images and 50,000 testing images of the ImageNet ISVRC 2012 dataset. Of the 135,000 pairs of images, 90\% are used as the training set and 10\% are used for validation and testing. The pipeline to collect this data is shown in Fig.~\ref{fig:datacollection}, which mainly consists of two steps: image capture and alignment.

%FIGURE moireDemo -----------------------------------------------------------------------

\paragraph{Image Capture}
Each reference image is enhanced with a black border and displayed at the centre of a computer screen (Fig.~\ref{fig:datacollection}(a)).
The reason to use black for the border is that we observe dark colours are least affected by the moir\'{e} effect. To increase the number of corner points that can be used during image alignment, we further extrude a black block from every edge of the black border. We then fill the rest of the screen outside the black border (and blocks) with pure white, which enables us to easily detect the black border in the captured images. We capture displayed images using a mobile phone (Fig.~\ref{fig:datacollection}(d)). During image acquisition, we randomly change the distance and angle between the mobile phone and the computer screen. Note that we require the black image borders to be always captured.

Detailed information of the phone models and the monitor screens is shown in Table~\ref{table:phoneModel} and Table~\ref{table:screenModel}, respectively. For each combination of phone model and screen, we collected 15,000 pairs of images. Thus, we collected $15,000 \times 9 = 135,000$ image pairs in total. Using different phone models as our capture devices ensures that moir\'{e} patterns are captured across different optical sensors while the diversity of display screens exhibits the difference in screen resolution.

\begin{table}[b]
\centering
\caption{Phone Model Specifications}
\label{table:phoneModel}
\begin{tabular}{|c|c|c|}
\midrule
\textbf{Manufacturer} & \textbf{Model}         & \textbf{Camera} \\ \midrule
APPLE                 & iPhone 6               & 8MP             \\ \midrule
SAMSUNG               & Galaxy S7 Edge         & 12MP            \\ \midrule
SONY                  & Xperia Z5 Premium Dual & 23MP            \\
\midrule
\end{tabular}
\end{table}

%\vspace{+3em}
\begin{table}[b]
\centering
\caption{Display Screen Specifications}
\label{table:screenModel}
\resizebox{.38\textwidth}{!}{\begin{tabular}{|c|c|c|c|}
\midrule
\textbf{Manufacturer} & \textbf{Model}         & \textbf{Resolution}  & \textbf{Size (inch)} \\ \midrule
APPLE                 & Macbook Pro Retina
& $2560 \times 1600$  & 13.3"          \\ \midrule
DELL               	  & U2410 LCD
& $1920 \times 1200$  & 24"          \\ \midrule
DELL                  & SE198WFP LCD
& $1280 \times 800 $  &   19"         \\ \midrule
\end{tabular}}
\end{table}
%------------------------------------------------------------------------------------

%FIGURE Data Collection -------------------------------------------------------------------
%------------------------------------------------------------------------------------

\paragraph{Image Alignment}
The prepared reference images and their corresponding captured images contaminated with moir\'{e} patterns have different resolutions and perspective distortions. To train our deep network in an end-to-end manner, we need to register them. 

In practice, we rely on the corners along the black image border to accomplish image alignment. Since we use a flat computer screen, the four corners of a captured image (excluding the blocks extruded from the border) lie on a plane. So do the four corners of the prepared reference image. Therefore, corresponding points in both the captured image and reference image are associated via a homography, which can be represented with a $3\times3$ projective matrix with 8 degrees of freedom. The four black blocks we attached to the image border increase the number of non-collinear corresponding points from 4 to 20, which can improve the registration precision. We use these 20 corners to compute the projective matrix and further align every pair of images.

\begin{figure}[t]
\begin{center}
\subfigure{
\includegraphics[width = 0.31\linewidth]{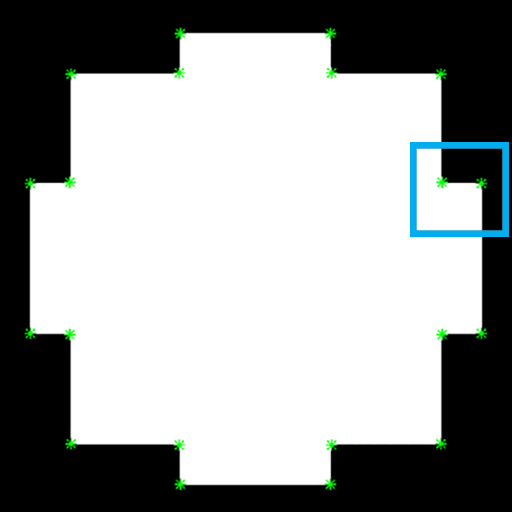}}
\hspace{-0.5em}
\subfigure{
\includegraphics[width = 0.31\linewidth]{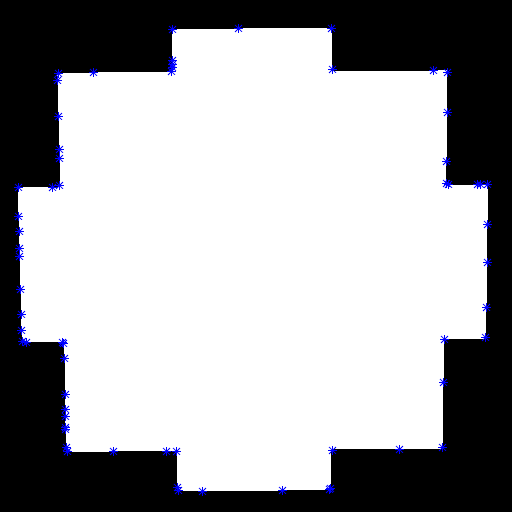}}
\hspace{-0.5em}
\subfigure{
\includegraphics[width = 0.31\linewidth]{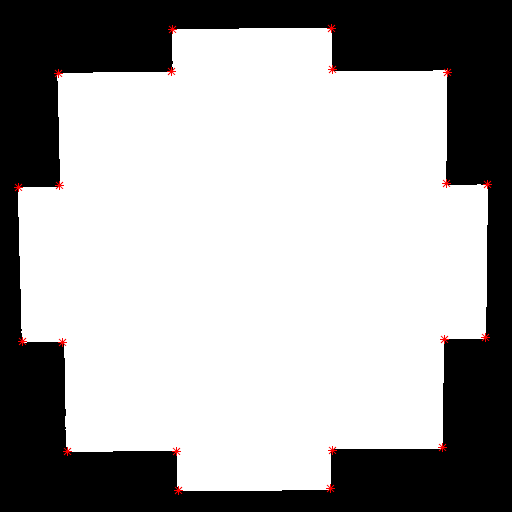}}
\end{center}
\vspace{-1.5em}
\addtocounter{subfigure}{-3}
\begin{center}
\subfigure[Detected corners of T  ]{
\includegraphics[width = 0.31\linewidth]{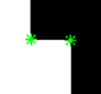}}
\hspace{-0.5em}
\subfigure[Detected corners of S]{
\includegraphics[width = 0.31\linewidth]{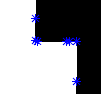}}
\hspace{-0.5em}
\subfigure[Cleaned corners of S]{
\includegraphics[width = 0.31\linewidth]{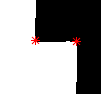}}
\end{center}

\caption{Corner Detection and Clearance.}
\label{fig:cornerdetection}
\end{figure}

To detect the corners, we convert the images into binary images and search for corners along the outermost boundary of the black image border. Traditional corner detection methods, such as the Harris corner detector~\cite{harris1988combined}, can faithfully detect all corners in a target image (Fig.~\ref{fig:cornerdetection}(a)). However, because of the presence of moir\'{e} artefacts, they fail to robustly find the 20 corresponding corners in the source image (Fig.~\ref{fig:cornerdetection}(b)), where certain edge pixels can be falsely detected as corners. 

To eliminate such false corners, we check the ratio between the numbers of black pixels and white pixels in a square neighbourhood around each detected corner. Since each corner forms a right angle, ideally, the ratio between the numbers of black and white pixels should be either $3$ or $1/3$. According to this observation, we filter out false corners, where the ratio between the numbers of black and white pixels in a square neighbourhood is clearly different from $3$ or $1/3$. In practice, we set the neighbourhood size to $11\times 11$. To remove duplicate corners, we set a minimum distance between two distinct corners. When the pairwise distances among two or more detected corners fall below this threshold, we only keep one of them. As shown in Fig.~\ref{fig:cornerdetection}(c), these twenty corners can be successfully detected. 

Finally, with the computed projective matrix, we can align every image pair.  The registration results are demonstrated in Fig.~\ref{fig:datacollection}(c) and ~\ref{fig:datacollection}(f).

%FIGURE Corner Detection cont.-------------------------------------------------------------------
%------------------------------------------------------------------------------------

\begin{figure}[t]
\begin{center}
 \includegraphics[width=0.7\linewidth]{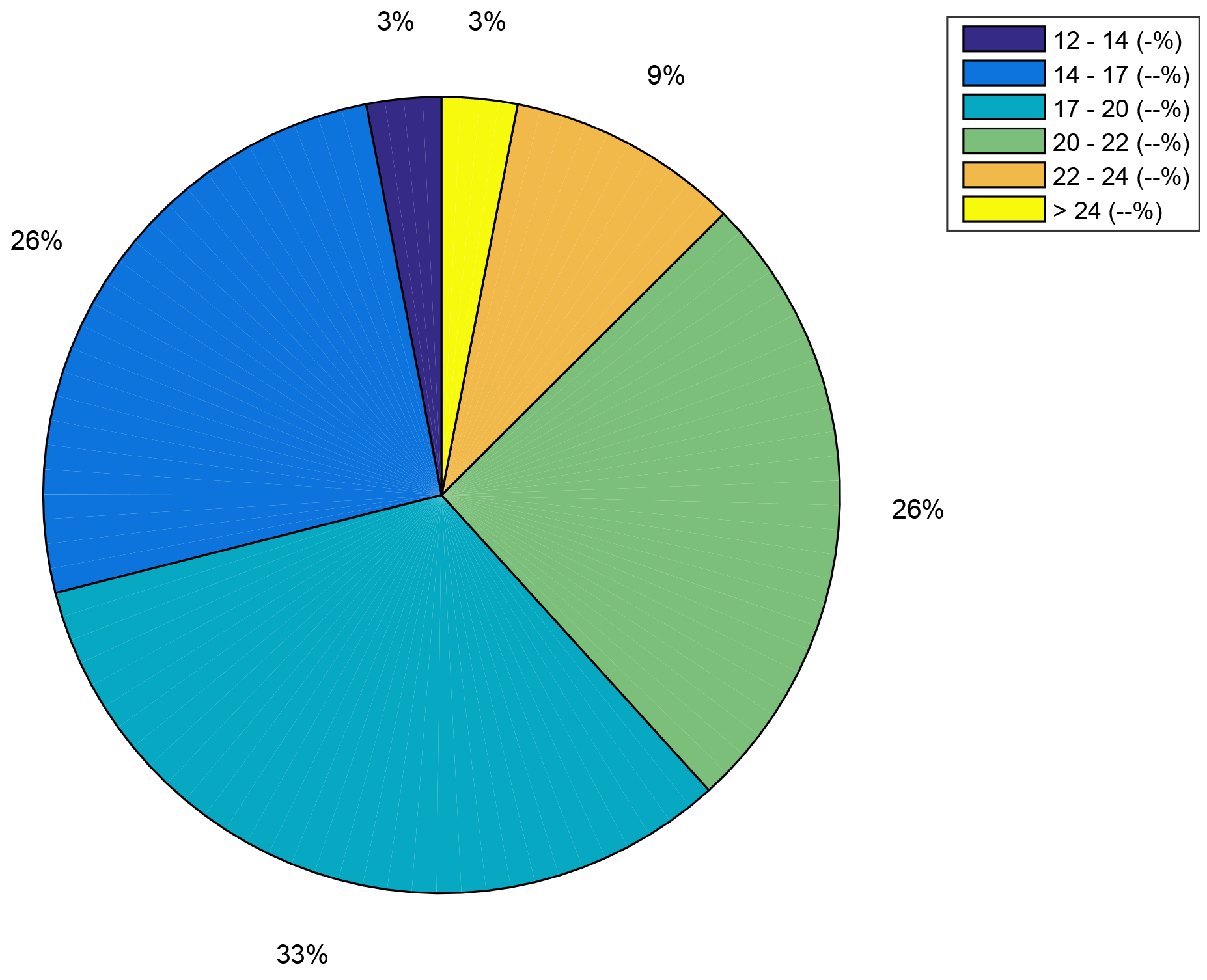}
\end{center}
 \caption{The quality distribution of moir\'{e} photos in the entire dataset. The quality of a moir\'{e} photo with respect to its corresponding reference image is measured using PSNR (dB).}
\label{fig:barInput}
\end{figure}

\paragraph{Automatic Verification} To automatically verify whether a registration result is correct or not, we measure the PSNR of the registered image pair and use a threshold $\eta$ to screen the PSNR value. In our experiments, we set $\eta = 12$. we have found that even images with the most severe moir\'{e} artefacts achieve PSNR values higher than 12dB while false registrations produce PSNR values lower than 10dB. The quality distribution of moir\'{e} photos in our dataset is shown in Fig.~\ref{fig:barInput}.

%FIGURE PSNRdemo -------------------------------------------------------------------
\begin{figure}[t]
\begin{center}
\subfigure[Moir\'{e}  19.9] {
\includegraphics[width = 0.24\linewidth]{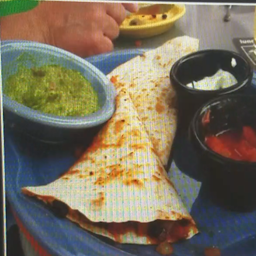}}
\hspace{-.8em}
\subfigure [GT ]{
\includegraphics[width = 0.24\linewidth]{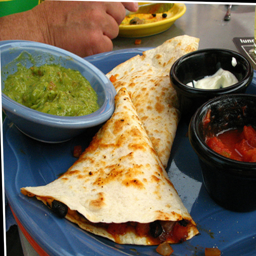}}
\hspace{-.8em}
\subfigure[Moir\'{e}  21.3]{
\includegraphics[width = 0.24\linewidth]{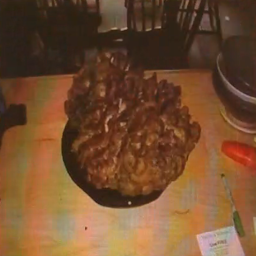}}
\hspace{-.8em}
\subfigure[GT ]{
\includegraphics[width = 0.24\linewidth]{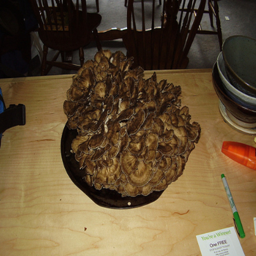}}
\end{center}
\caption{PSNR cannot fully reflect the degree of moir\'{e} patterns. An image corrupted by visually more severe moir\'{e} patterns can have higher PSNR.}
\label{fig:PSNRdemo}
\end{figure}

However, note that PSNR cannot fully reflect the severity of the moir\'{e} effect. As shown in Fig.~\ref{fig:PSNRdemo}, an image corrupted by a visually more severe moir\'{e} pattern actually achieves a higher PSNR. This is perhaps because the colour bands in a moir\'{e} pattern do not significantly affect PSNR even though they are visually disturbing and easily noticeable.

%FIGURE plot Bar Chart input -----------------------------------------------------------------------
%-----------------------------------------------------------------

\paragraph{Setup}
During image acquisition, images are displayed on the screen consecutively. Each reference image stays on the screen for 0.3 seconds. We use a mobile phone to record a video of the consecutively displayed images. Frames from the captured video are then retrieved as images contaminated with moir\'{e} patterns.

%_---------------------------------------------------------------
\section{Model Understanding and Implementation}
\subsection{Insights Behind Our Network Design}
\begin{figure}[t]
\begin{center}
\subfigure{\includegraphics[width=0.24\linewidth]{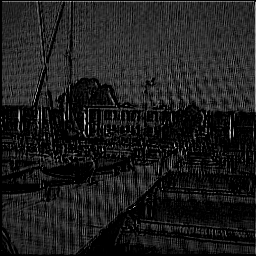}}
\subfigure{\includegraphics[width=0.24\linewidth]{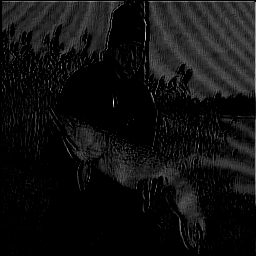}}
\subfigure{\includegraphics[width=0.24\linewidth]{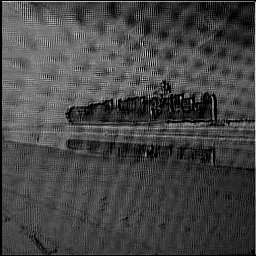}}
\subfigure{\includegraphics[width=0.24\linewidth]{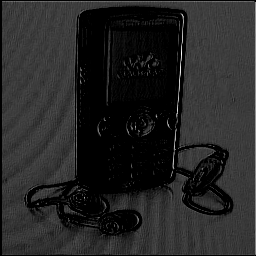}}
\end{center}
\addtocounter{subfigure}{-3}
\vspace{-1.5em}
\begin{center}
\subfigure{\includegraphics[width=0.24\linewidth]{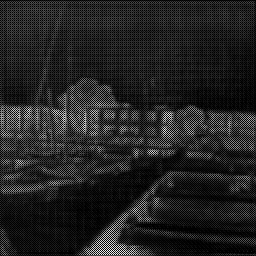}}
\subfigure{\includegraphics[width=0.24\linewidth]{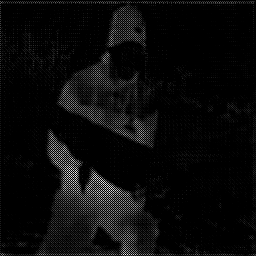}}
\subfigure{\includegraphics[width=0.24\linewidth]{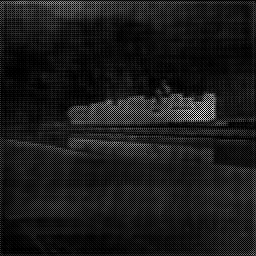}}
\subfigure{\includegraphics[width=0.24\linewidth]{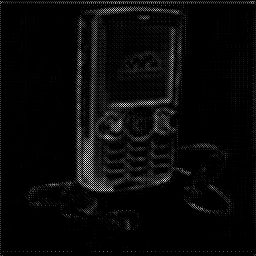}}
\end{center}
 \caption{Visualisation of U-Net feature maps. (Top) Feature maps produced by a layer A closer to the input image. (Bottom) Feature maps produced by a deeper layer B. Layer A is skipped connected to layer B.}
\label{fig:unetFeature}
\end{figure}

Moir\'{e} patterns span a wide range in both spatial and frequency domains. Therefore, we conceive a multi-resolution architecture, which has convolutional layers with multi-scale receptive fields, to tackle this problem. At the beginning, we experimented with U-Net~\cite{ronneberger2015u} with skip connections. Skip connections have been proven to be effective in high-level vision tasks, such as image recognition and semantic segmentation. However, when tackling low-level vision problems, including super-resolution, denoising and deblurring, many approaches can produce state-of-the-art results without skip connections, such as VDSR, DnCNN and PyramidCNN. In high-level vision problems, the information from high-resolution layers close to the input image is useful for the additional clues they introduce. Different from other tasks making use of networks with skip connections, moir\'{e} photos and their corresponding ground-truth images can differ dramatically, and thus, skip connections are not powerful enough to model such differences. In addition, the layer closer to the input image in a skip connection contains serious moir\'{e} artefacts, as shown in the top row of Fig.~\ref{fig:unetFeature}, while the feature maps produced by the deeper layer are relatively moir\'{e}-free.  As a result, directly using high-frequency details from a layer closer to the input image would likely introduce artefacts in the final result.

\begin{figure*}[htb!]
\begin{center}
\subfigure{\includegraphics[width=0.16\linewidth]{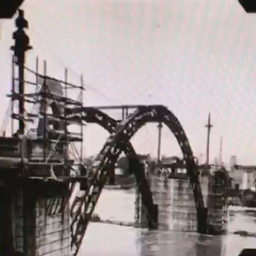}}
\subfigure{\includegraphics[width=0.16\linewidth]{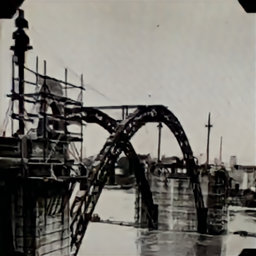}}
\subfigure{\includegraphics[width=0.32\linewidth]{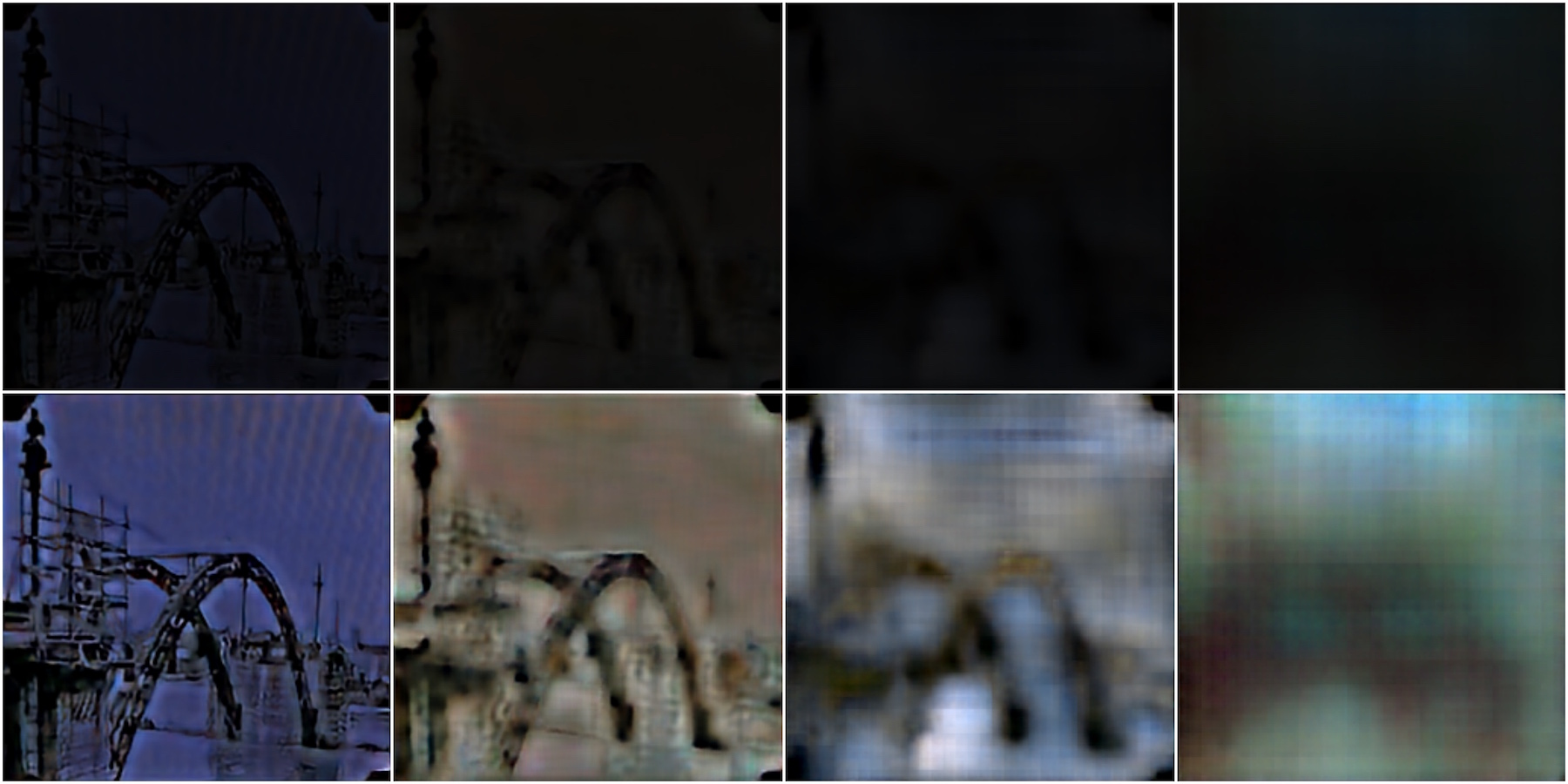}}
\subfigure{\includegraphics[width=0.16\linewidth]{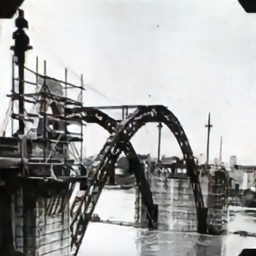}}
\end{center}
\addtocounter{subfigure}{-4}
\vspace{-1.2em}
\begin{center}
\subfigure[Input]{\includegraphics[width=0.16\linewidth]{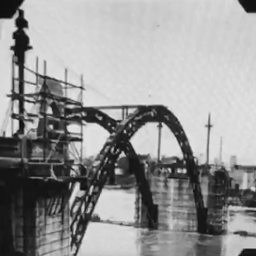}}
\subfigure[Finest Scale]{\includegraphics[width=0.16\linewidth]{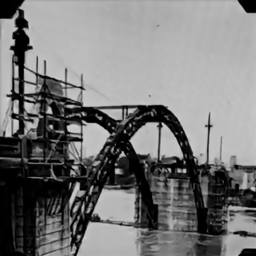}}
\subfigure[Scale 2 to 5]{\includegraphics[width=0.32\linewidth]{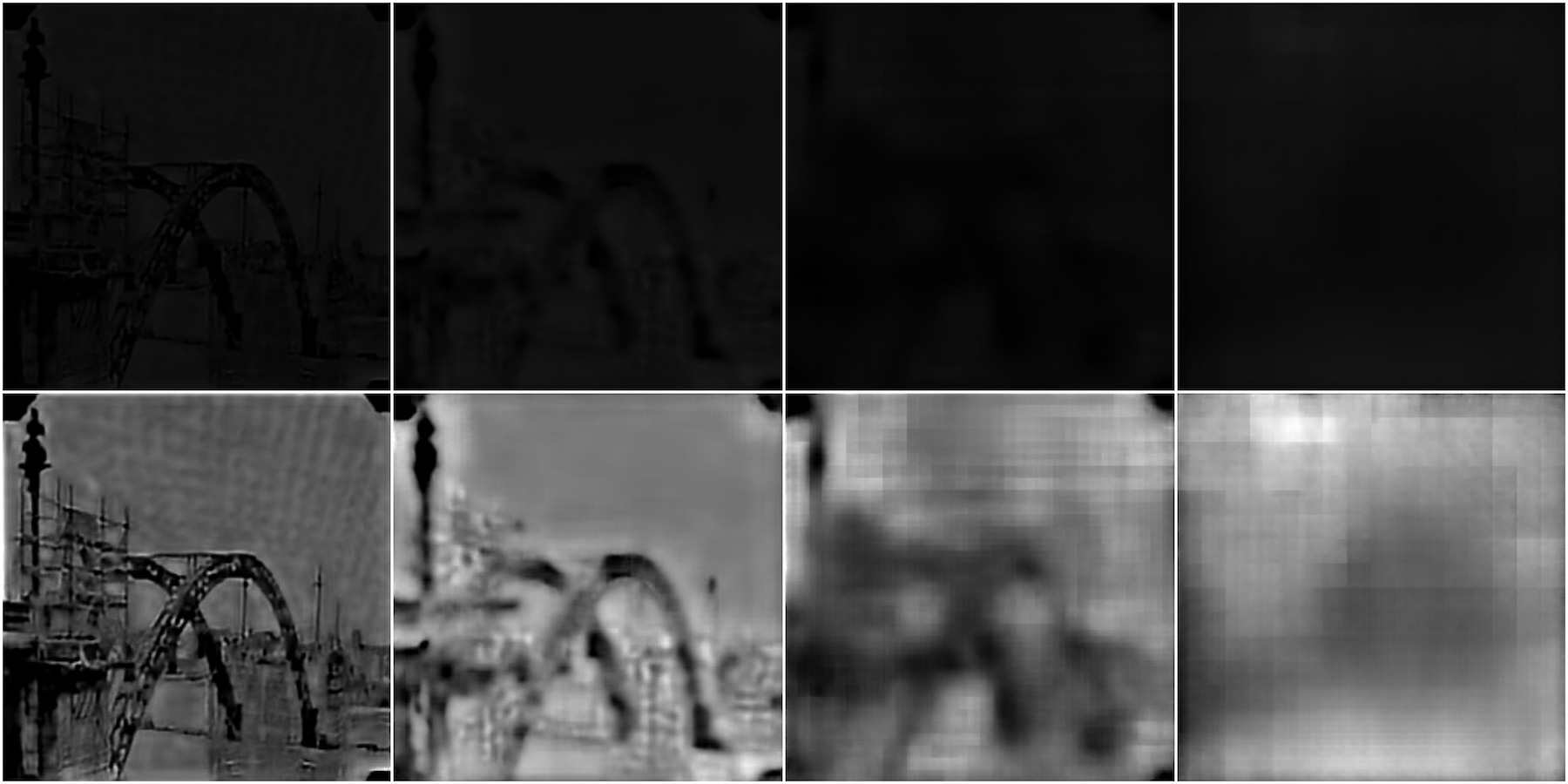}}
\subfigure[Output]{\includegraphics[width=0.16\linewidth]{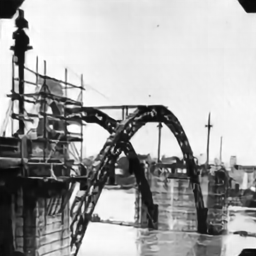}}
\end{center}
 \caption{Visualisation of the 3-channel feature maps produced by different branches on a ``grayscale-like" RGB image and its corresponding pure grayscale image. For each input, the top row in (c) shows the intermediate images produced from the second to the fifth network branch, and the bottom row shows the same images with amplified intensity.}
\label{fig:scaleDemo1}
\end{figure*}

PyramidCNN~\cite{nah2016deep} also adopts a multi-resolution architecture for deblurring. In their architecture, an input image is first downsampled to $k$ resolutions linearly and then network branches for different resolutions are trained simultaneously. For the task of deblurring, coarser level output guides the training process of finer level network branches. But for moir\'{e} pattern removal, the output from coarser levels is not completely free of moir\'{e} artefacts, which tend to make finer levels maintain such artefacts.

To achieve better performance, we embed a multi-resolution pyramid in our network architecture. In contrast to traditional image pyramids built with linear filtering, the image pyramid in our architecture is actually built with nonlinear filtering because nonlinear activation always follows each convolutional layer. The nonlinearity in our pyramid allows the network to perform more effectively during downsampling. More importantly, in our network, each resolution is associated with a network branch with six stacked convolutional layers maintaining the same resolution. Such network branches are capable of performing sophisticated nonlinear transformations (such as removing moir\'{e} artefacts within a specific frequency band), and are more powerful than skip connections in U-Net.

\subsection{A Detailed Study on Our Proposed Model}
To show the advantage of the proposed model, we attempt to test different variants. Model specifications  are given as follows:
\begin{itemize}
\item V\_Concate 	(27.12dB): replacing the sum operation with concatenation. To be specific, we concatenate the 32 feature maps from each scale, and append two convolutional layers after the concatenated feature maps. Each of these convolutional layers has 32 channels and $3 \times 3$ kernels.
\item V\_Skip    		(26.36dB): in each scale, skip connecting the second downsampling layer to the last convolutional layer before the upsampling layers.
\item V\_C32    		(25.52dB): replacing all the 64-channel convolution filters with 32 channel convolutional filters.
\item V\_B123  		(25.28dB): using branch 1, 2 and 3 only.
\item V\_B135  		(26.04dB): using branch 1, 3 and 5 only.
\item V\_B15    		(25.52dB): using branch 1 and 5 only.
\end{itemize}
We will demonstrate later that although V\_Concate achieves a higher PSNR score on the test data, it produces worse visual results than our proposed network. Adding skip connections cannot further improve the performance of the proposed model while the other variants degrade the performance.

%Table. PSNRcompare -------------------------------------------------------------
\begin{table*}[t]
\centering
\caption{A quantitative comparison among participating methods on our test set with different metrics. Our method clearly outperforms the other methods.}
\label{table:PSNRcompare}
\resizebox{.99\textwidth}{!}{ \begin{tabular}{@{}|c|c|c|c|c|c|c|c|c|c|c|@{}}
\toprule
 & Corrected Input & RTV~\protect{\cite{xu2012structure}}  & SDF~\protect{\cite{ham2015}} & IRCNN~\protect{\cite{zhang2017learning}} & DnCNN~\protect{\cite{zhang2017beyond}} & VDSR~\protect{\cite{kim2016accurate}} &  PyramidCNN~\protect{\cite{nah2016deep}} & U-Net~\protect{\cite{ronneberger2015u}} & V\_Concate & Our method           \\ \midrule
PSNR Mean (dB)       		& 20.30        & 20.67    &    20.88     &   21.01   &  24.54  	  &  24.68  & 25.39   & 26.49   & 27.12    &26.77 \\ \midrule
PSNR Gain (dB)         		&       -         & 0.37      &    0.58       &   0.71     &  4.24    	  &   4.38   & 5.09     &  6.19    & 7.09      & 6.47 \\ \midrule
Ave Error ($\times 10^{-3}$)      &  34      	  & 31    	  &    30    	      &  28.32    &  5.82    	  &    5.74  & 4.83     & 3.81     & 3.36      & 3.62  \\ \midrule
SSIM~\cite{ssim}       					& 0.738      & -    &    -      &   -   &    0.834     &  0.837   & 0.859   & 0.864  & 0.878    & 0.871 \\ \midrule
FSIM~\cite{fsim}      					& 0.869      & -    &    -      &   -   &    0.901     &  0.902   & 0.909   & 0.912  & 0.922    & 0.914 \\ \bottomrule
\end{tabular}}
\end{table*}
%--------------------------------------------------------------

\subsection{Grayscale Moir\'{e} Artefacts}
To verify that our model can remove moir\'{e} patterns rather than the unnatural colours, we convert the RGB dataset to a grayscale one and retrain the network. The average PSNR, SSIM and FSIM on the grayscale testing set are 27.26, 0.852, and 0.910, respectively, indicating that our model is able to deal with moir\'{e} patterns regardless of the colour information. Intermediate images produced from different branches on a test RGB image that is close to a grayscale one as well as those produced on its corresponding grayscale image are demonstrated in Fig.~\ref{fig:scaleDemo1}. 

\subsection{Implementation} We have fully implemented our proposed deep multiresolution network using CAFFE on an NVIDIA Geforce 1080 GPU. The entire training process takes 3 days on average. We use a mini-batch size of 8, start with learning rate $0.0001$, set the weight decay to $0.00001$, and minimize the loss function using Adam~\cite{kingma2014adam}. We have found that the training process could not converge properly with a higher learning rate. As the training process proceeds, we reduce the learning rate by a factor of 10 when the loss on a validation set stops decreasing. In all the experiments in this paper, we set the patch size $p\times p$ to 256 $\times$ 256. The network weights are randomly initialised using a Gaussian with a zero mean and a standard deviation equal to 0.01. The bias in each neuron is initialised to 0.

%------------------------------------------------------------------------------------
\section{Comparison and Discussion}
In this section, we experimentally analyse our method's capability in improving image quality and removing moir\'{e} artefacts. Since we are not aware of any existing methods that solve exactly the same problem, we compare our method against state-of-the-art methods in related image restoration problems, including image denoising, deblurring, super-resolution and texture removal. We choose VDSR~\cite{kim2016accurate} as a representative from image super-resolution algorithms, DnCNN~\cite{zhang2017beyond} and IRCNN~\cite{zhang2017learning} from the latest image denoising methods, and RTV~\cite{xu2012structure} and SDF~\cite{ham2015} among texture removal techniques. For that a subset of the moir\'{e} photos in our dataset has a certain degree of blurriness and that deblurring techniques can reconstruct high-frequency details, we also add two latest learning based image deblurring techniques, multi-scale pyramidCNN~\cite{nah2016deep} and IRCNN~\cite{zhang2017learning}, for comparison. Moreover, since we adopt a hierarchical network architecture, we also compare our network with U-Net~\cite{ronneberger2015u}, an effective neural network for image segmentation.

To perform a fair comparison, we tune the parameters of the methods we compare against so that they reach the optimal performance on our dataset. When a method only has a small number of tuneable parameters, we tune those parameters to make the method achieve the lowest average error on our test set. When a method has a large number of parameters, such as learning based methods, we retrain the model in the method using our training set.

Even though descreening methods aim at removing a different and simpler moir\'{e} effect that occurs in scanned copies of printed documents and images, they are certainly relevant. Since such methods are relatively mature and have been integrated into commercial software, we choose to compare with the descreening function in Photoshop.

%FIGURE plotLoss --------------------------------------------
\begin{figure}[t]
\begin{center}
 \includegraphics[width=0.99\linewidth]{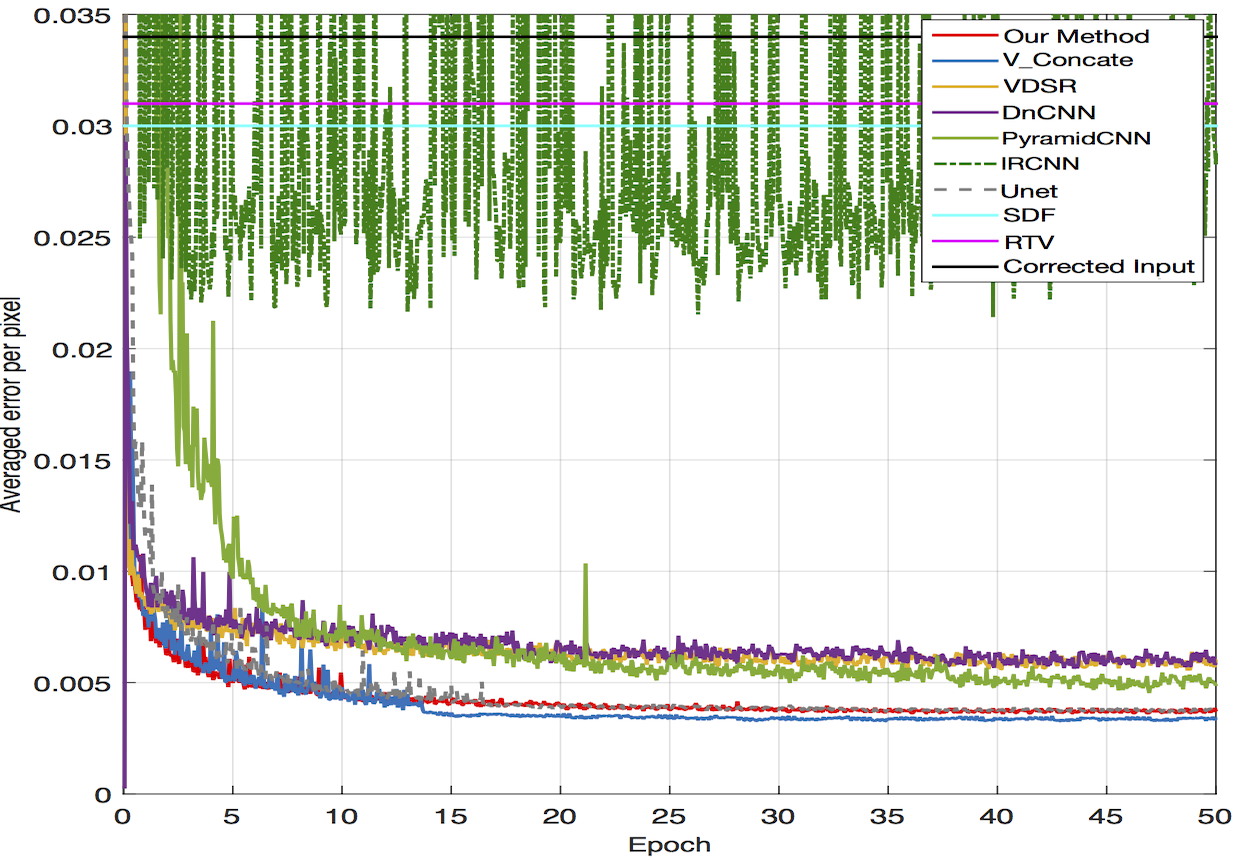}
\end{center}
 \caption{Average pixel-wise MSE error of various methods vs. the number of epochs.}
\label{fig:plotLoss}
\end{figure}
%--------------------------------------------------------------

%FIGURE compare -------------------------------------------------------------------
\begin{figure*}
\begin{center}
\subfigure[ Input 17.7]{
\includegraphics[width = 0.14\linewidth]{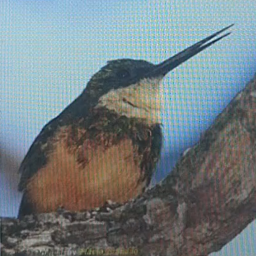}}
\hspace{-0.5em}
\subfigure [RTV 17.6]{
\includegraphics[width = 0.14\linewidth]{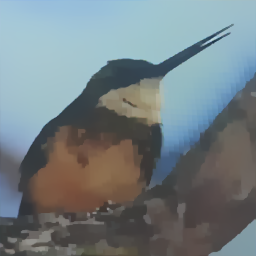}}
\hspace{-0.5em}
\subfigure [SDF 17.5]{
\includegraphics[width = 0.14\linewidth]{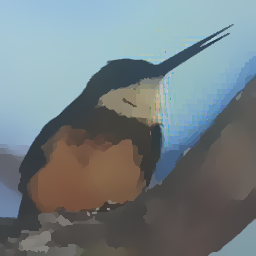}}
\hspace{-0.5em}
\subfigure [Descreen 17.3]{
\includegraphics[width = 0.14\linewidth]{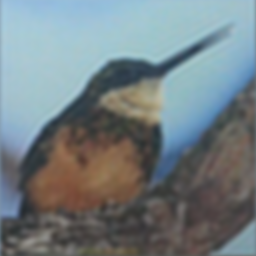}}
\hspace{-0.5em}
\subfigure [IRCNN 18.8]{
\includegraphics[width = 0.14\linewidth]{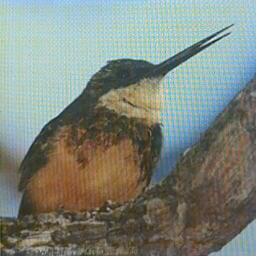}}
\subfigure [U-Net 22.7]{
\includegraphics[width = 0.14\linewidth]{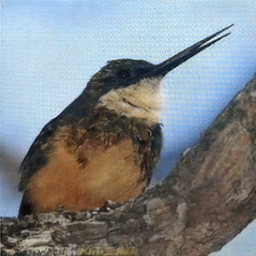}}
\end{center}
\vspace{-1.3em}
\begin{center}
\subfigure [VDSR 22.9]{
\includegraphics[width = 0.14\linewidth]{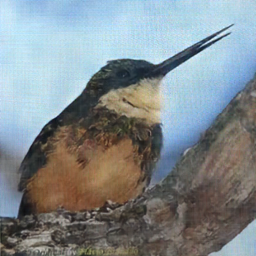}}
\hspace{-0.5em}
\subfigure [DnCNN 22.2]{
\includegraphics[width = 0.14\linewidth]{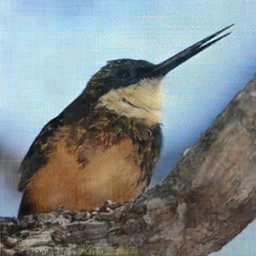}}
\hspace{-0.5em}
\subfigure [PyramidCNN 22.2]{
\includegraphics[width = 0.14\linewidth]{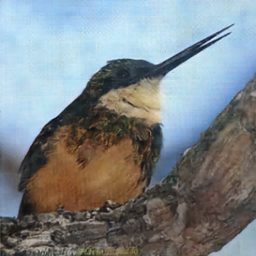}}
\hspace{-0.5em}
\subfigure[V\_Concate 24.9] {
\includegraphics[width = 0.14\linewidth]{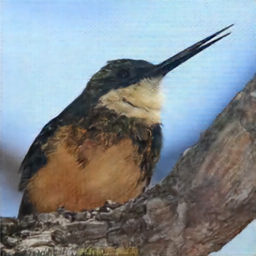}}
\hspace{-0.5em}
\subfigure [Our method 24.6]{
\includegraphics[width = 0.14\linewidth]{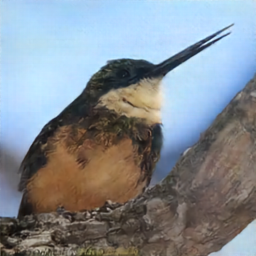}}
\hspace{-0.5em}
\subfigure [Ground Truth]{
\includegraphics[width = 0.14\linewidth]{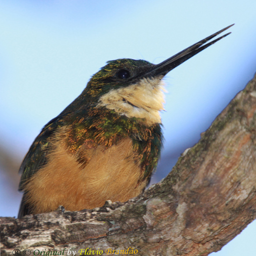}}
\end{center}
\vspace{-.5em}

\addtocounter{subfigure}{-12}
\begin{center}
\subfigure[ Input 21.8]{
\includegraphics[width = 0.14\linewidth]{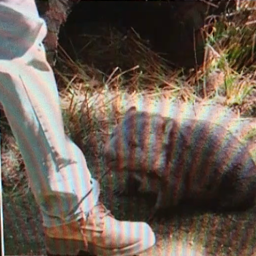}}
\hspace{-0.5em}
\subfigure [RTV 21.1]{
\includegraphics[width = 0.14\linewidth]{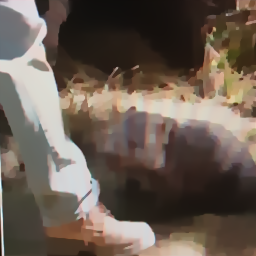}}
\hspace{-0.5em}
\subfigure [SDF 21.4]{
\includegraphics[width = 0.14\linewidth]{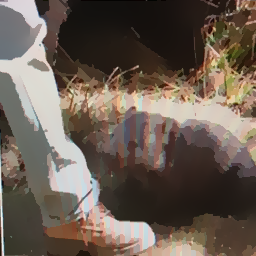}}
\hspace{-0.5em}
\subfigure [Descreen 20.0]{
\includegraphics[width = 0.14\linewidth]{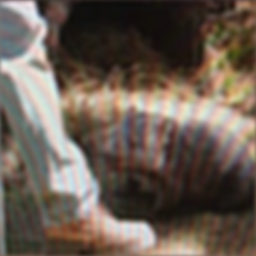}}
\hspace{-0.5em}
\subfigure [IRCNN 22.1]{
\includegraphics[width = 0.14\linewidth]{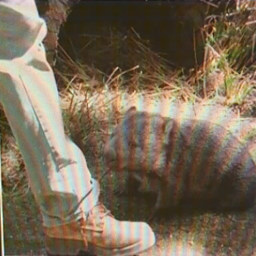}}
\hspace{-0.5em}
\subfigure [U-Net 27.2]{
\includegraphics[width = 0.14\linewidth]{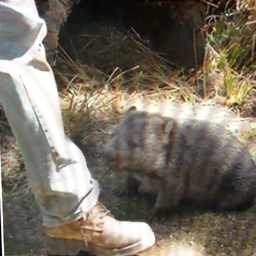}}
\end{center}
\vspace{-1.3em}

\begin{center}
\subfigure [VDSR 22.5]{
\includegraphics[width = 0.14\linewidth]{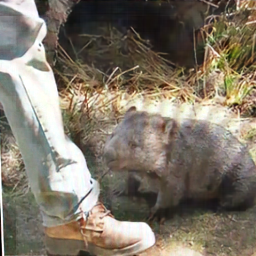}}
\hspace{-0.5em}
\subfigure [DnCNN 23.1]{
\includegraphics[width = 0.14\linewidth]{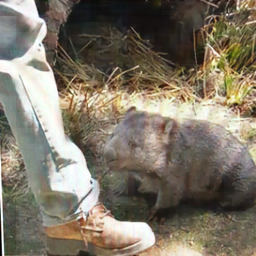}}
\hspace{-0.5em}
\subfigure [PyramidCNN 24.6]{
\includegraphics[width = 0.14\linewidth]{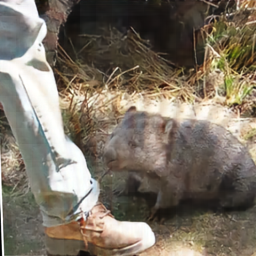}}
\hspace{-0.5em}
\subfigure [V\_Concate 28.3]{
\includegraphics[width = 0.14\linewidth]{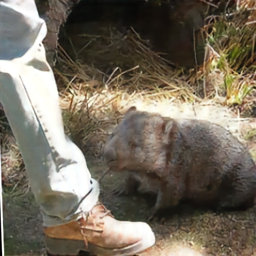}}
\hspace{-0.5em}
\subfigure [Our method 27.6]{
\includegraphics[width = 0.14\linewidth]{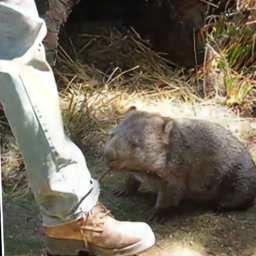}}
\hspace{-0.5em}
\subfigure [Ground Truth]{
\includegraphics[width = 0.14\linewidth]{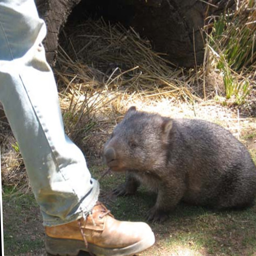}}
\end{center}

\caption{Comparison between our multiresolution deep network and other state-of-the-art methods for image restoration, including Photoshop Descreen,
IRCNN~\protect{\cite{zhang2017learning}},
U-Net~\cite{ronneberger2015u},		
VDSR~\protect{\cite{kim2016accurate}},
DnCNN~\protect{\cite{zhang2017beyond}},
pyramidCNN~\protect{\cite{nah2016deep}},
RTV~\protect{\cite{xu2012structure}} and
SDF~\protect{\cite{ham2015}}.}
\label{fig:compareALL}
\end{figure*}
%------------------------------------------------------------------------------------

\subsection{Quantitative Comparison}
In Fig.~\ref{fig:plotLoss} and Table~\ref{table:PSNRcompare}, we demonstrate the quantitative performance of different methods on our test set. Since the contaminated image and the reference image within the same pair have different average intensity levels due to multiple reasons, including the brightness of the computer screen and the intensity response curve of the camera during image acquisition, that are mostly irrelevant to the moir\'{e} effect, we decided to factor out the differences in average intensity by adjusting the average intensity of the contaminated image to be the same as that of the reference image (Corrected Input). As shown, our method and the variant of our model, V\_Concate, outperform all other methods participating in the comparison on all performance measures, including PSNR, SSIM~\cite{ssim} and FSIM~\cite{fsim}.
As the parameters for descreening in Photoshop have to be adjusted manually for each image, we cannot show the average performance on the entire test set. However, we will qualitatively compare it with our method in the next section.

Effective as a super-resolution method, VDSR~\cite{kim2016accurate} delivers a reasonable performance but is unable to fully handle the complex moir\'{e} effect. Using a configuration with a large receptive field, the denoising network (DnCNN) in \cite{zhang2017beyond} has a similar performance as VDSR~\cite{kim2016accurate}. Both VDSR and DnCNN adopt a flat CNN architecture that maintains the same resolution across all layers. Nonetheless, both of them have been clearly outperformed by our multiresolution network.

By defining a denoising prior with dilated convolutions, IRCNN~\cite{zhang2017learning} outperforms state-of-the-art methods in pixel-wise image restoration tasks. However, it performs poorly on our dataset and its training process can hardly converge on our training set. After modifying IRCNN by interleaving ordinary convolutions and dilated convolutions, we obtain a revised model called IRCNN-IL. The convergence issue is resolved in the revised model but its performance is still not satisfactory. The PSNR, SSIM and FSIM achieved by IRCNN-IL are 21.55, 0.744, and 0.870, respectively.
In theory, the noise IRCNN aims to deal with is completely different from the moir\'{e} patterns we attempt to remove. A noisy image is commonly modelled as the result of an additive process, which adds noise to the original signal, but a moir\'{e} pattern is a phenomenon caused by light interference, which is a different and much more complicated process. Dilated kernels can remove additive noises but might be insufficient to remove complex moir\'{e} patterns. Due to the different underlying mechanisms of image noises and  moir\'{e} patterns, one cannot be certain that IRCNN is effective for restoring moir\'{e} photos.

Nah {\em et al.}~\shortcite{nah2016deep} deblur images bottom up using a multiresolution Gaussian pyramid. It first deblurs an image in $1/2^i$ resolution, then in $1/2^{i-1}$ resolution and finally in the full resolution. The multiresolution architecture helps to produce acceptable results. However, unlike our multiresolution pyramid generated from trainable nonlinear filters (convolutional kernels), their pyramid is generated using the fixed Gaussian filter, which is linear. As shown in Fig.~\ref{fig:plotLoss} and Table~\ref{table:PSNRcompare}, our network architecture delivers clearly better performance.

Among all the methods, U-Net~\cite{ronneberger2015u} achieves a numerical performance closest to our method. However, we found that even though U-Net produces good statistics, it delivers relatively poor visual results, which will be demonstrated in visual comparisons. Likewise, V\_Concate produces the highest score on all metrics but its ability in visually removing moir\'{e} patterns is less than the original model.

Texture removal techniques, RTV~\cite{xu2012structure} and SDF~\cite{ham2015}, are useful in preserving important image structures while eliminating small repetitive textural details. But image features at a similar scale of texture elements would be removed as well. In our context, these techniques are used for removing moir\'{e} patterns, and they give a poor performance on this task. The difficulty in setting an appropriate texture kernel size could be the main reason because a large smoothing and texture kernel would over-smooth the image while a small kernel would not be able to remove low-frequency large-scale moir\'{e} artefacts.

\subsection{Visual Comparisons}
We visually compare results from our method against those from other state-of-the-art methods in Fig.~\ref{fig:compareALL}. Additional visual comparisons can be found in the supplemental materials. Note that the input images are all from the test set. From these comparisons, we have the following observations. RTV~\cite{xu2012structure} and SDF~\cite{ham2015} remove small-scale texture features which typically have higher frequencies than moir\'{e} patterns. Descreening in Photoshop over-smoothes the input image. Among deep learning based methods, IRCNN~\cite{zhang2017learning} is unable to remove moir\'{e} patterns at all even though its network has been re-trained using our training set. Meanwhile, VDSR\cite{kim2016accurate}, PyramidCNN~\cite{nah2016deep}, and DnCNN~\cite{zhang2017beyond} have a better performance. However, colour distortion is still noticeable in their results.

\begin{figure}[t]
\begin{center}
\subfigure[Input 16.1]{
\includegraphics[width = 0.31\linewidth]{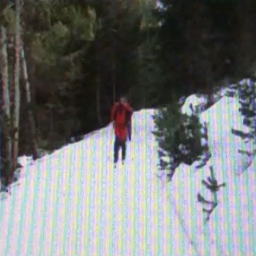}}
\subfigure[U-Net~\cite{ronneberger2015u} 26.2]{
\includegraphics[width = 0.31\linewidth]{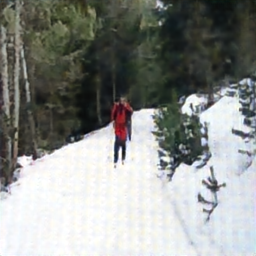}}
\subfigure[Our method 26.0 ]{
\includegraphics[width = 0.31\linewidth]{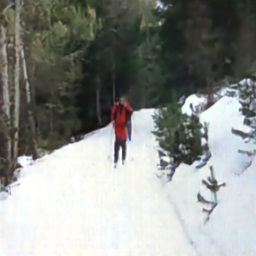}}
\end{center}

\caption{Another example in which U-Net~\cite{ronneberger2015u} produces a higher PSNR score but a worse moir\'{e} removal effect. }
\label{fig:anotherUnetExample}
\end{figure}

Except for our methods, U-Net~\cite{ronneberger2015u} achieves the highest scores of all quality measures. But more moir\'{e} artefacts remain in its results than in the results of VDSR\cite{kim2016accurate} and DnCNN~\cite{zhang2017beyond}. As we have stated earlier, even though a quality measure, such as PSNR, can measure the overall image quality, it cannot precisely measure the effectiveness in moir\'{e} pattern removal.
We show an example in Fig.~\ref{fig:anotherUnetExample} and the supplemental materials that U-Net~\cite{ronneberger2015u} produces higher PSNRs but worse visual results. Our method has the most powerful network architecture and produces output images closest to the ground-truth reference images.

\begin{figure*}
\begin{center}
\subfigure{
\includegraphics[width = 0.14\linewidth]{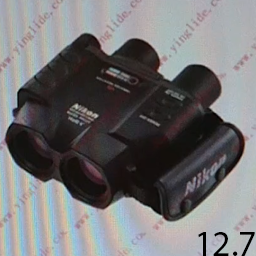}}
\subfigure{
\includegraphics[width = 0.14\linewidth]{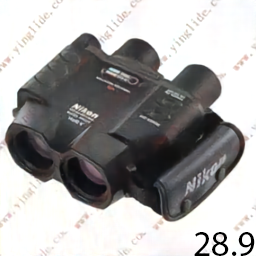}}
\subfigure{
\includegraphics[width = 0.14\linewidth]{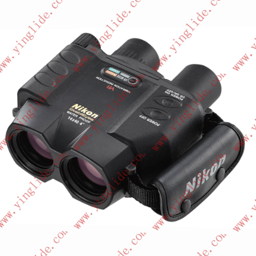}}
\subfigure{
\includegraphics[width = 0.14\linewidth]{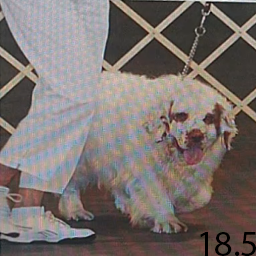}}
\subfigure{
\includegraphics[width = 0.14\linewidth]{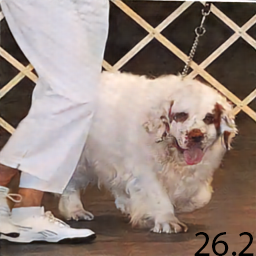}}
\subfigure{
\includegraphics[width = 0.14\linewidth]{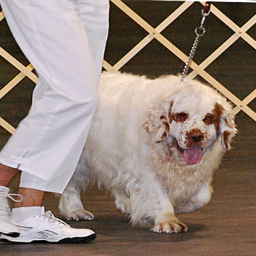}}
\end{center}
\addtocounter{subfigure}{-6}
\vspace{-1em}

\begin{center}
\subfigure{
\includegraphics[width = 0.14\linewidth]{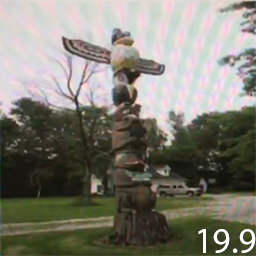}}
\subfigure{
\includegraphics[width = 0.14\linewidth]{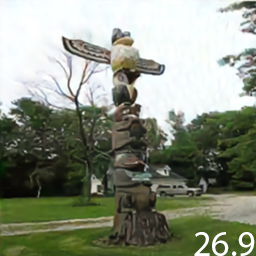}}
\subfigure{
\includegraphics[width = 0.14\linewidth]{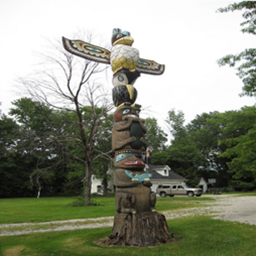}}
\subfigure{
\includegraphics[width = 0.14\linewidth]{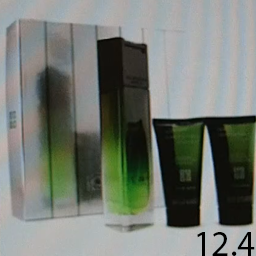}}
\subfigure{
\includegraphics[width = 0.14\linewidth]{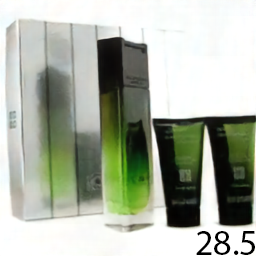}}
\subfigure{
\includegraphics[width = 0.14\linewidth]{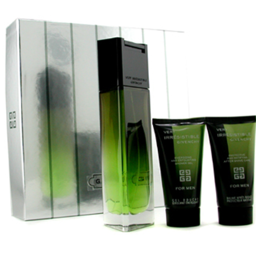}}
\end{center}
\addtocounter{subfigure}{-6}
\vspace{-1em}

\begin{center}
\subfigure{
\includegraphics[width = 0.14\linewidth]{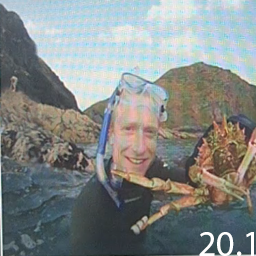}}
\subfigure{
\includegraphics[width = 0.14\linewidth]{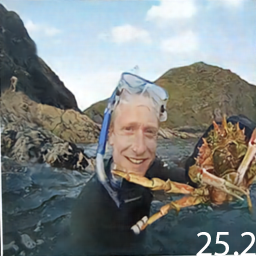}}
\subfigure{
\includegraphics[width = 0.14\linewidth]{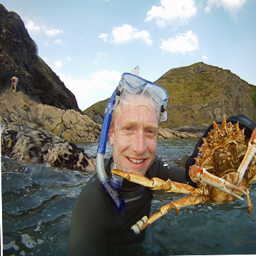}}
\subfigure{
\includegraphics[width = 0.14\linewidth]{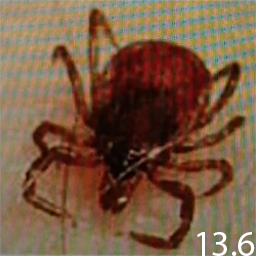}}
\subfigure{
\includegraphics[width = 0.14\linewidth]{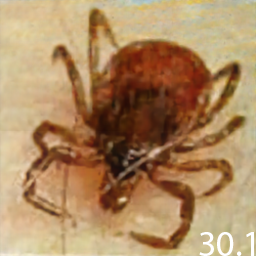}}
\subfigure{
\includegraphics[width = 0.14\linewidth]{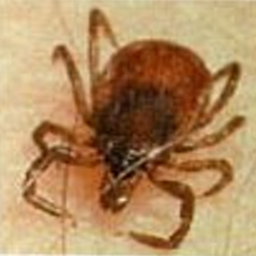}}
\end{center}
\addtocounter{subfigure}{-6}
\vspace{-1em}

\begin{center}
\subfigure{
\includegraphics[width = 0.14\linewidth]{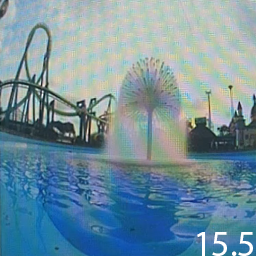}}
\subfigure{
\includegraphics[width = 0.14\linewidth]{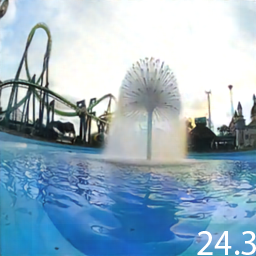}}
\subfigure{
\includegraphics[width = 0.14\linewidth]{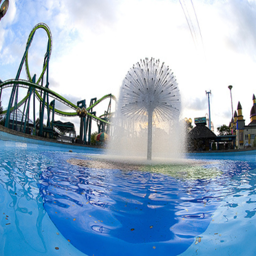}}
\subfigure{
\includegraphics[width = 0.14\linewidth]{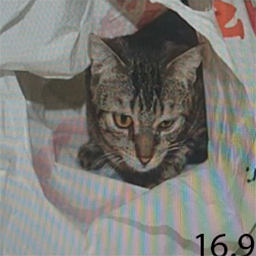}}
\subfigure{
\includegraphics[width = 0.14\linewidth]{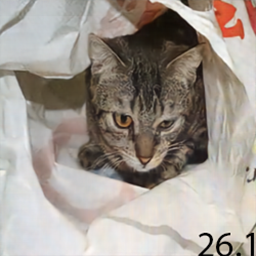}}
\subfigure{
\includegraphics[width = 0.14\linewidth]{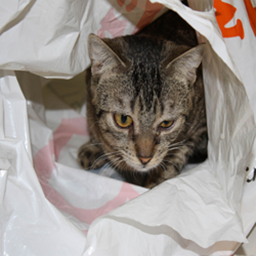}}
\end{center}
\addtocounter{subfigure}{-6}
\vspace{-1em}

\begin{center}
\subfigure{
\includegraphics[width = 0.14\linewidth]{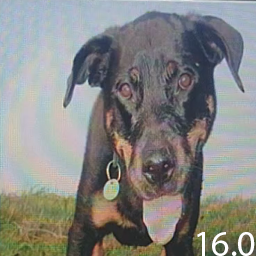}}
\subfigure{
\includegraphics[width = 0.14\linewidth]{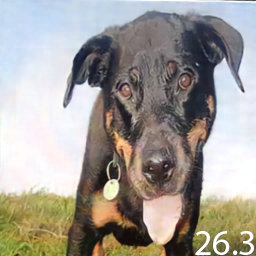}}
\subfigure{
\includegraphics[width = 0.14\linewidth]{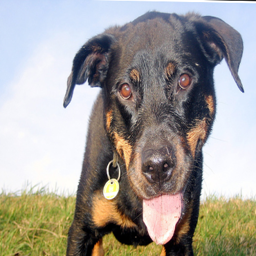}}
\subfigure{
\includegraphics[width = 0.14\linewidth]{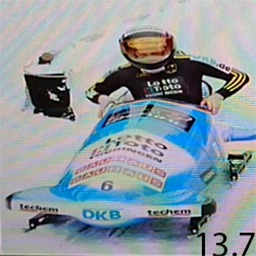}}
\subfigure{
\includegraphics[width = 0.14\linewidth]{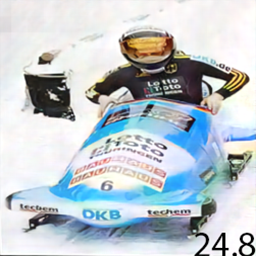}}
\subfigure{
\includegraphics[width = 0.14\linewidth]{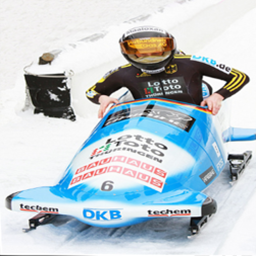}}
\end{center}
\addtocounter{subfigure}{-6}
\vspace{-1em}

\begin{center}
\subfigure{
\includegraphics[width = 0.14\linewidth]{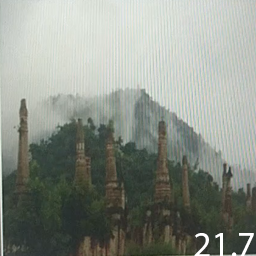}}
\subfigure{
\includegraphics[width = 0.14\linewidth]{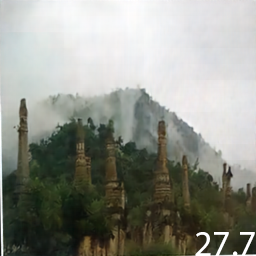}}
\subfigure{
\includegraphics[width = 0.14\linewidth]{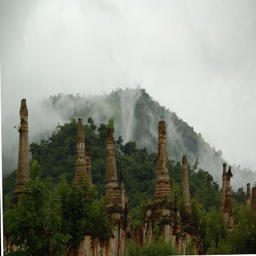}}
\subfigure{
\includegraphics[width = 0.14\linewidth]{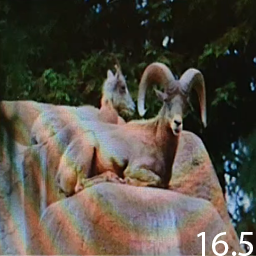}}
\subfigure{
\includegraphics[width = 0.14\linewidth]{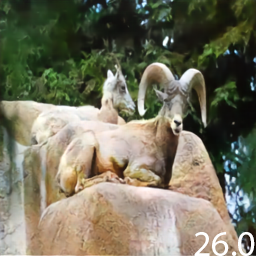}}
\subfigure{
\includegraphics[width = 0.14\linewidth]{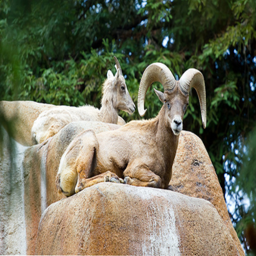}}
\end{center}
\addtocounter{subfigure}{-6}
\vspace{-1em}

\begin{center}
\subfigure{
\includegraphics[width = 0.14\linewidth]{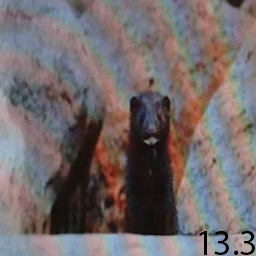}}
\subfigure{
\includegraphics[width = 0.14\linewidth]{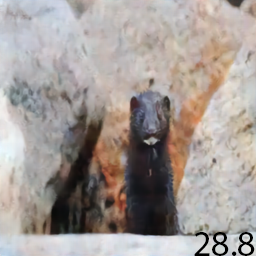}}
\subfigure{
\includegraphics[width = 0.14\linewidth]{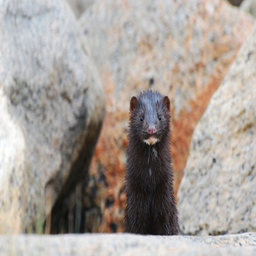}}
\subfigure{
\includegraphics[width = 0.14\linewidth]{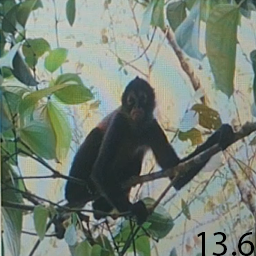}}
\subfigure{
\includegraphics[width = 0.14\linewidth]{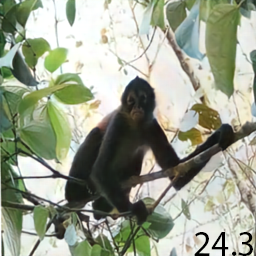}}
\subfigure{
\includegraphics[width = 0.14\linewidth]{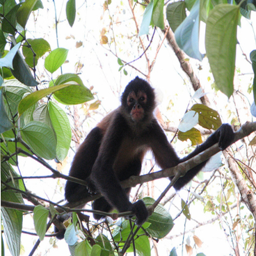}}
\end{center}
\addtocounter{subfigure}{-6}
\vspace{-1em}

\begin{center}
\subfigure[Input]{
\includegraphics[width = 0.14\linewidth]{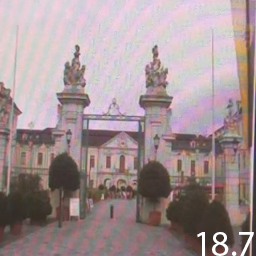}}
\subfigure[Our Result]{
\includegraphics[width = 0.14\linewidth]{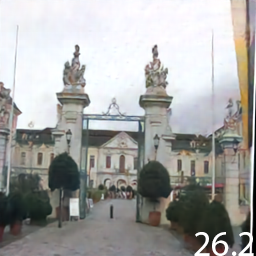}}
\subfigure[Ground Truth]{
\includegraphics[width = 0.14\linewidth]{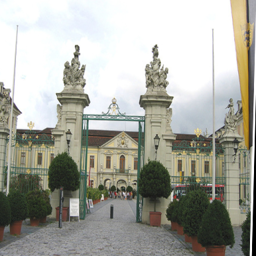}}
\subfigure[Input]{
\includegraphics[width = 0.14\linewidth]{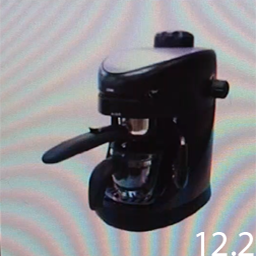}}
\subfigure[Our Result]{
\includegraphics[width = 0.14\linewidth]{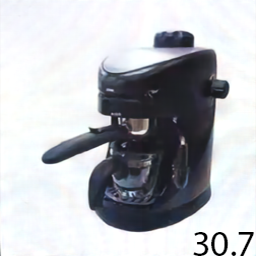}}
\subfigure[Ground Truth]{
\includegraphics[width = 0.14\linewidth]{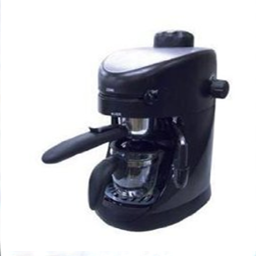}}
\end{center}

\vspace{-.5em}

\caption{Input images contaminated with different types of moir\'{e} patterns and their corresponding cleaned results from our proposed method. In this figure, we intentionally show some brighter images, where moir\'{e} patterns are more noticeable.}
\label{fig:result1}
\end{figure*}

Additional visual results from our method are shown in Fig.~\ref{fig:result1}, where the input images exhibit a variety of different moir\'{e} patterns.

%%FIGURE another compare with Unet -------------------------------------------------------------------

\subsection{The Number of Variables}
As shown in Table.~\ref{table:numPara}, the number of variables in our method is in the same order as U-Net and PyramidCNN while our proposed network outperforms both of them qualitatively and quantitatively. Variants of our model, V\_B15 and V\_C32, have a similar number of parameters as VDSR and DnCNN, however produce higher PSNR scores.

\begin{table}[thb!]
\centering
\caption{The number of variables in learning based approaches ($\times 10^5$).}
\label{table:numPara}
\resizebox{0.48\textwidth}{!}{\begin{tabular}{|c|c|c|c|c|c|}
\hline
                                    & V\_B123    & V\_B15 	& V\_C32    	& V\_Concate & Our method    	\\ \hline
\# var 			   & 9.28           & 7.42   	&4.11		& 16.14         	 &    15.44    		\\ \hline
                                    & IRCNN-IL     & VDSR      & DnCNN 	& PyramidCNN & U-Net    		 	 \\ \hline
\# var  			   &  3.35   	      & 6.67  	& 7.04    		&14.15    		& 24.62      		 \\ \hline
\end{tabular}}
\end{table}

\subsection{User Study}
\begin{figure}[htb!]
\begin{center}
 \includegraphics[width=0.99\linewidth]{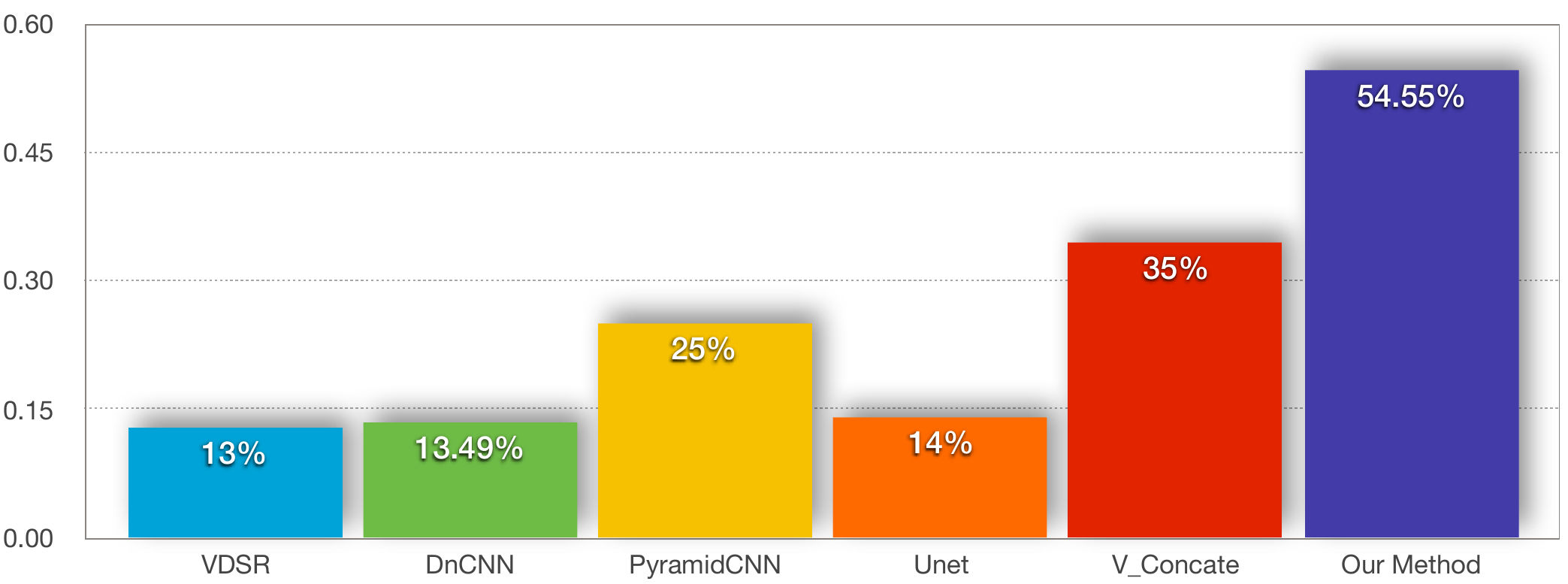}
\end{center}
 \caption{User study on moir\'{e} pattern restoration.}
\label{fig:userStudychart}
\end{figure}

Due to the limitation of image metrics in measuring moir\'{e} artefacts, we have also conducted a user study to compare different methods, which includes 20 questions. Each question consists of six randomly ordered results, generated by VDSR, DnCNN, PyramidCNN, U-Net, V\_Concate and our method,  on a randomly selected test image. 60 participants have to choose 1 to 2 images that they perceive most appealing and comfortable. After averaging the votes from all the 20 questions, we obtain the statistics in Fig.~\ref{fig:userStudychart}. It is clear that the proposed model is more preferable to the human visual system, although U-Net and V\_Concate achieve high scores under certain numerical image quality measures.

%---------------------------------------------------------------
\section{Model Versatility}

\subsection{Cross-Data Evaluation}
We quantitatively measure our model versatility by training and testing on data collected with different phone models or digital monitors. We perform three experiments, including testing on images taken with an iPhone on a Mac 2560 screen, with a SamSung S7 on a Dell 1920 monitor, and with a Sony Z5 on a Dell 1280 display, respectively. Note that in each experiment, the test data is excluded during training process. The performance is demonstrated in Table.~\ref{table:crossV}. Though the performance is not as good as before, our model can still produce reasonable results. We also observe that the quality improvement by our model is most noticeable when the input (moir\'{e}) images are in low quality, such as the images captured with the Sony Z5 on the DELL 1280 screen.

\begin{table}[htb!]
\centering
\caption{Cross-data evaluation.}
\label{table:crossV}
\resizebox{0.48\textwidth}{!}{\begin{tabular}{|c|c|c|c|c|c|c|}
\hline
\multirow{2}{*}{Test Data\textbackslash Metrics} & \multicolumn{2}{c|}{PSNR} & \multicolumn{2}{c|}{SSIM} & \multicolumn{2}{c|}{FSIM} \\ \cline{2-7}
                                                & Input       & Result      & Input       & Result      & Input       & Result      \\ \hline
iPhone\_Mac2560                                 & 23.09       & 25.18       & 0.840       & 0.862       & 0.914       & 0.930       \\ \hline
SamSung\_Dell1920                               & 18.34       & 20.84       & 0.594       & 0.636       & 0.833       & 0.870       \\ \hline
Sony\_Dell1280                                       &  16.33           &     23.28        &    0.706         &  0.822           &   0.856          &  0.898           \\ \hline
\end{tabular}}
\end{table}

\begin{figure}[htb!]
\begin{center}
\subfigure[HUAWEI P9  ]{
\includegraphics[width = 0.24\linewidth]{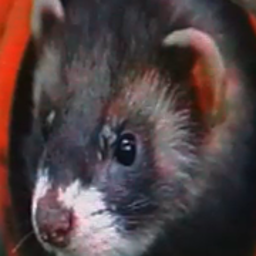}}
\hspace{-.8em}
\subfigure[Our result ]{
\includegraphics[width = 0.24\linewidth]{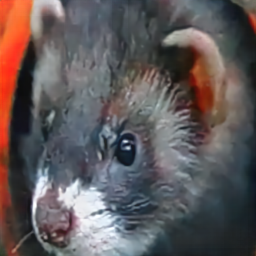}}
\hspace{-.8em}
\subfigure[HUAWEI P9  ]{
\includegraphics[width = 0.24\linewidth]{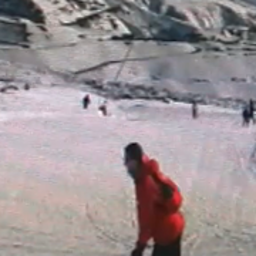}}
\hspace{-.8em}
\subfigure[Our result ]{
\includegraphics[width = 0.24\linewidth]{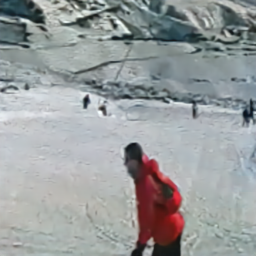}}
\end{center}
\caption{Restoration of moir\'{e} photos taken with HUAWEI P9. Our model is not fine-tuned for this phone model.}
\label{fig:otherphone}
\end{figure}

\paragraph{Test on phone model HUAWEI P9} Though the camera sensors in different phone models are different, the underlying reason for the formation of moir\'{e} patterns is similar on different phones. To test the versatility of our network, we run our network directly on moir\'{e} photos captured by another phone model, that is not used in collecting our dataset, HUAWEI P9. Decent results have been achieved, as shown in Fig.~\ref{fig:otherphone} and the supplemental materials. This indicates that our trained network can be used for removing moir\'{e} patterns in images captured by other phone models.

%FIGURE otherphone -------------------------------------------------------------------

\subsection{Restore Partial Moir\'{e} Photos}
\begin{figure}[htb!]
\begin{center}
\subfigure[Input]{\includegraphics[width=0.24\linewidth]{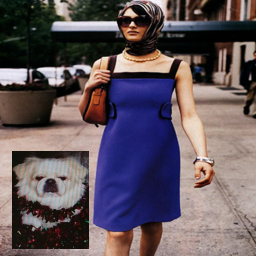}}
\subfigure[Our Result]{\includegraphics[width=0.24\linewidth]{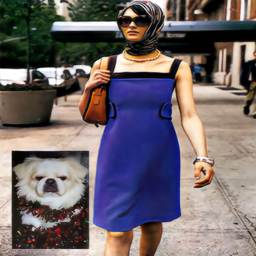}}
\subfigure[Input]{\includegraphics[width=0.24\linewidth]{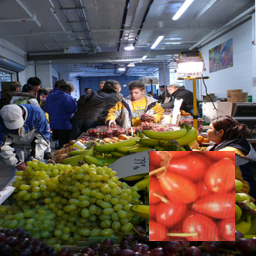}}
\subfigure[Our result]{\includegraphics[width=0.24\linewidth]{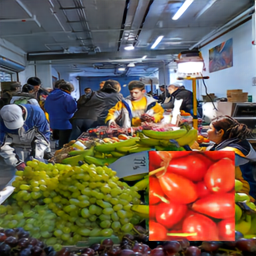}}
\end{center}
\caption{Test on synthesised images contaminated by moir\'{e} patterns in a small region. }
\label{fig:partialmoire}
\end{figure}

\paragraph{Synthesised moir\'{e} images} Moir\'{e} patterns on an image can be spatially varying, strong in a region and weak in another region. Under extreme conditions, moir\'{e} patterns can only appear in part of an image. In Fig.~\ref{fig:partialmoire}, we show our results on synthesised partial moir\'{e} images, where only a small portion of the image contains moir\'{e} artefacts.

\begin{figure}[htb!]
\begin{center}
\subfigure{\includegraphics[width=0.3\linewidth]{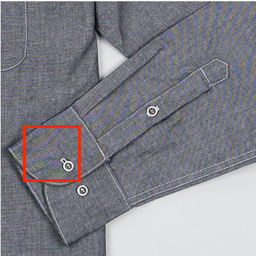}}
\subfigure{\includegraphics[width=0.3\linewidth]{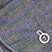}}
%\subfigure{\includegraphics[width=0.3\linewidth]{images/onlineResults/image_10_demoire_246}}
\subfigure{\includegraphics[width=0.3\linewidth]{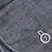}}
\end{center}
\addtocounter{subfigure}{-3}
\vspace{-1.2em}
\begin{center}
\subfigure[Input]{\includegraphics[width=0.3\linewidth]{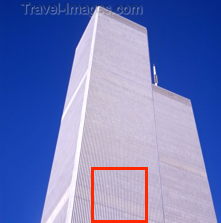}}
\subfigure[Input-CloseUp]{\includegraphics[width=0.3\linewidth]{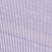}}
%\subfigure{\includegraphics[width=0.3\linewidth]{images/onlineResults/image_7_demoire_263}}
\subfigure[Our result]{\includegraphics[width=0.3\linewidth]{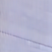}}
\end{center}
 \caption{Reduce moir\'{e} artefacts on Internet images without fine-tuning. Image courtesy @Fstoppers user Peter House and @Travel-Images.com user A.Bartel, respectively.}
\label{fig:online2}
\end{figure}

\paragraph{Real world moir\'{e} patterns not caused by display} When searching the Internet for ``moir\'{e} photos", we find that moir\'{e} patterns most commonly appear on fine repetitive patterns,  such as textile textures on clothes and buildings. In Fig.~\ref{fig:online2}, we show the results of directly applying our trained model without fine-tuning on Internet images damaged by moir\'{e} artefacts.  Though the moir\'{e} is caused by the repetition of the fine patterns rather than digital display, our model is able to reduce such moir\'{e} patterns as well.

%--------------------------------------------------------------------------------------------------------------------------------------------
\section{Limitations }
\begin{figure}[thb!]
\begin{center}
\subfigure[Input ]{
\includegraphics[width = 0.295\linewidth]{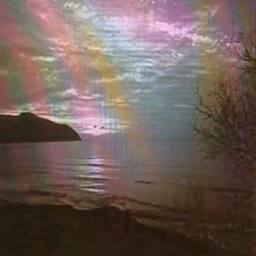}}
\subfigure[Our method ]{
\includegraphics[width = 0.295\linewidth]{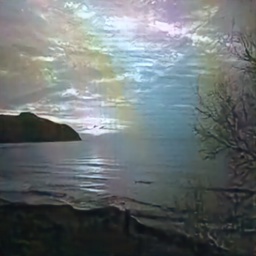}}
\subfigure[Ground truth ]{
\includegraphics[width = 0.295\linewidth]{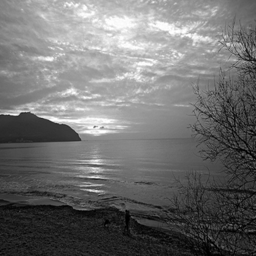}}
\end{center}

\caption{A failure example. }
\label{fig:failure}
\end{figure}

When a moir\'{e} pattern exhibits very severe large-scale coloured bands, our method might not be able to infer the uncontaminated image correctly. We show a failure case in Fig.~\ref{fig:failure}.

Another limitation is that our model could not clearly reduce blurriness in the input images. Note that other baseline algorithms, including the image deblurring model PyramidCNN, are not able to resolve it either (Fig.~\ref{fig:compareALL}). We believe that such blurriness is introduced into a subset of acquired photos in our dataset because of multiple reasons, including motion blur due to the movement of the camera during image acquisition, the imperfect image alignment during pre-processing, and the damaged high-frequency components caused by high-frequency moir\'{e} patterns. Although our algorithm can faithfully detect all 20 corner points, moir\'{e} patterns can interfere with their exact localisation, giving rise to imperfect alignment.

%FIGURE Failure -------------------------------------------------------------------

%-----------------------------------------------------------------------
%FIGURE Result1
%%--------------------------------------------------------------------

%--------------------------------------------------------------
\section{Conclusion and Future Work}
To conclude, we presented a novel multiresolution fully convolutional network for automatically removing moir\'{e} patterns from photos as well as created a large-scale benchmark with $100,000^+$ image pairs to evaluate moir\'{e} pattern removal algorithms. Although a moir\'{e} pattern can span over a wide range of frequencies, our proposed network is able to remove moir\'{e} artefacts within every frequency band thanks to the nonlinear multiresolution analysis of the moir\'{e} photos. We believe that people would like to use their mobile phones to record content on screens for more reasons than expected, such as convenience, simplicity, and efficiency. The proposed method and the collected large-scale benchmark together provide a decent solution to the moir\'{e} photo restoration problem. 

In the future, we would like to explore different categories of moir\'{e} patterns and improve our method so that it can eliminate moir\'{e} artefacts according to their category labels. Moreover, it will be interesting to investigate the existence of an indicator that can better describe the level of moir\'{e} artefacts and guide the training process. We also plan to keep expanding our dataset by adding more examples under different shooting conditions and for different types of device screens. We believe that with a larger dataset, our method can produce even better results. 

%------------------------------------------------------------------------------------

\section*{Acknowledgements.} 
This work was partially supported by Hong Kong Research Grants Council under General Research Funds (HKU17209714).

%% use section* for acknowledgment
%\section*{Acknowledgment}

\ifCLASSOPTIONcaptionsoff
  \newpage
\fi



%\bibliographystyle{IEEEtran}

\bibliographystyle{IEEEtran}
\bibliography{demoireCNN}

\end{document}